\documentclass[conference]{IEEEtran}
\ifCLASSINFOpdf
\else
\fi
\usepackage{amsmath,amssymb,amsfonts}
\usepackage{algorithmic}
\usepackage{graphicx}
\usepackage{textcomp}
\usepackage{xcolor}
\def\BibTeX{{\rm B\kern-.05em{\sc i\kern-.025em b}\kern-.08em
    T\kern-.1667em\lower.7ex\hbox{E}\kern-.125emX}}
\usepackage{multirow}
\usepackage{algorithm} % 排版算法
\usepackage{stfloats}
\usepackage{balance}
\makeatletter
\let\NAT@parse\undefined
\makeatother
\usepackage{hyperref}
\usepackage[inkscapelatex=false]{svg}
\usepackage{subfigure}
\usepackage{threeparttable}
\usepackage{color, colortbl}

\begin{document}

\title{\LARGE \bf Rethinking Urban Mobility Prediction: A Super-Multivariate Time Series Forecasting Approach \\
}

\author{Jinguo Cheng$^{1}$, Ke Li$^{1}$, Yuxuan Liang$^{2}$, Lijun Sun$^{3}$, Junchi Yan$^{4}$ and Yuankai Wu$^{1\dagger}$\footnote{*Corresponding author}% <-this % stops a space
\\
{$^{1}$Sichuan University.}
{$^{2}$The Hong Kong University of Science and Technology (Guangzhou).}
\\
{$^{3}$McGill University.}
{$^{4}$Shanghai Jiao Tong University and Shanghai AI Lab.}}

\maketitle

%\title{Long-term High-resolution Traffic Flow Forecasting}

%\title{Grid-based Traffic Flow is Super Multivariate Time Series}



\begin{abstract}
Long-term urban mobility predictions play a crucial role in the effective management of urban facilities and services. Conventionally, urban mobility data has been structured as spatiotemporal videos, treating longitude and latitude grids as fundamental pixels. Consequently, video prediction methods, relying on Convolutional Neural Networks (CNNs) and Vision Transformers (ViTs), have been instrumental in this domain. In our research, we introduce a fresh perspective on urban mobility prediction. Instead of oversimplifying urban mobility data as traditional video data, we regard it as a complex multivariate time series. This perspective involves treating the time-varying values of each grid in each channel as individual time series, necessitating a thorough examination of temporal dynamics, cross-variable correlations, and frequency-domain insights for precise and reliable predictions. To address this challenge, we present the Super-Multivariate Urban Mobility Transformer (SUMformer), which utilizes a specially designed attention mechanism to calculate temporal and cross-variable correlations and reduce computational costs stemming from a large number of time series. SUMformer also employs low-frequency filters to extract essential information for long-term predictions. Furthermore, SUMformer is structured with a temporal patch merge mechanism, forming a hierarchical framework that enables the capture of multi-scale correlations. Consequently, it excels in urban mobility pattern modeling and long-term prediction, outperforming current state-of-the-art methods across three real-world datasets.
 %As a result, it excels in regional mobility pattern modeling and long-term prediction, surpassing current state-of-the-art methods by 10\%. Our approach provides a fresh and promising outlook on the field of urban mobility forecasting.
\end{abstract}


% \fi
\IEEEpeerreviewmaketitle

\section{Introduction}

In the realm of urban mobility computing, a diverse array of spatiotemporal data exists, encompassing different organizational structures, scales, and modes. These data are characterized by their dynamic nature, evolving continuously across both time and space. Among the prominent forms of urban dynamic spatiotemporal data are point-based~\cite{wang2020deep}, graph-based~\cite{jin2023spatio}, and grid-based data~\cite{bigdataset}. Grid-based data, in particular, involve the division of urban areas into grids based on latitude and longitude coordinates. Each grid contains a wealth of attributes for the current spatiotemporal slot, including latitude and longitude coordinate ranges, points of interest, cumulative in/out vehicle counts (in/out traffic flow)~\cite{zhang2017deep}, and various other relevant information~\cite{liu2018attentive,lin2020preserving}. Forecasting grid-based data is crucial as it serves as a foundational framework for spatial analysis and modeling, enabling the assessment, prediction, and management of various urban phenomena, spanning from congestion hotspots to land use dynamics.

Traditionally, the practice of structuring grid-based mobility data in a video format $(T, C, H, W)$ has naturally emerged due to its alignment with the inherent characteristics of the data. Here, $T$ corresponds to the number of time points, $C$ represents the number of attributes, and $H$ and $W$ indicate the latitude and longitude dimensions of the urban area. In recent years, there have been remarkable advancements in deep video prediction techniques, with Convolutional Neural Networks (CNNs)~\cite{krizhevsky2012imagenet} and more recent Vision Transformers (ViTs)~\cite{dosovitskiy2020image, arnab2021vivit} serving as their core components. This has resulted in the extensive utilization of deep video prediction methods for the prediction of grid-based mobility data. Moreover, it is important to note that grid-based mobility data such as TaxiBJ also serves as a common dataset for evaluating video prediction models~\cite{simvp, tang2023swinlstm, guen2020disentangling, yu2019efficient}.
% \begin{figure}[htp]
% \centering    
% {\includesvg[width=0.4\textwidth]{traffic_flow_correlation3.svg}}
% \caption{Three types of correlation: different gird in the same channel, different channel in the same gird, different gird in different channel}
% \label{correlation}
% \end{figure}
\begin{figure*}
    \centering
    \subfigure[Conventional Embedding Approaches in CNNs and ViTs.]{ 
    \label{fig1b}
    \includegraphics[width=0.38\textwidth]{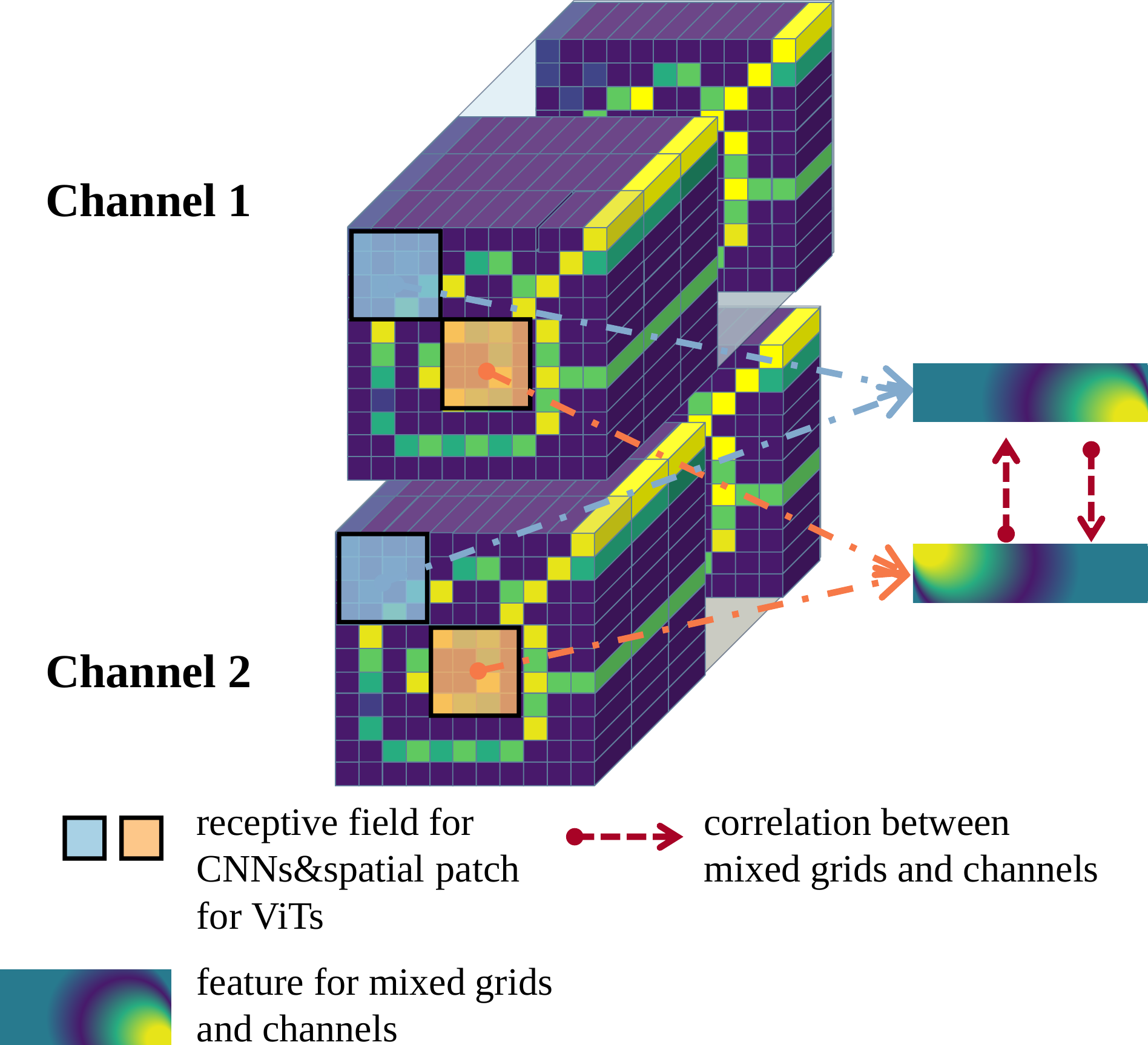}}
    \hspace{5pt}
    \subfigure[A Super-Multivariate Perspective on Grid-Based Urban Mobility Data.]{
    \label{fig1a}
    \includegraphics[width=0.45\textwidth]{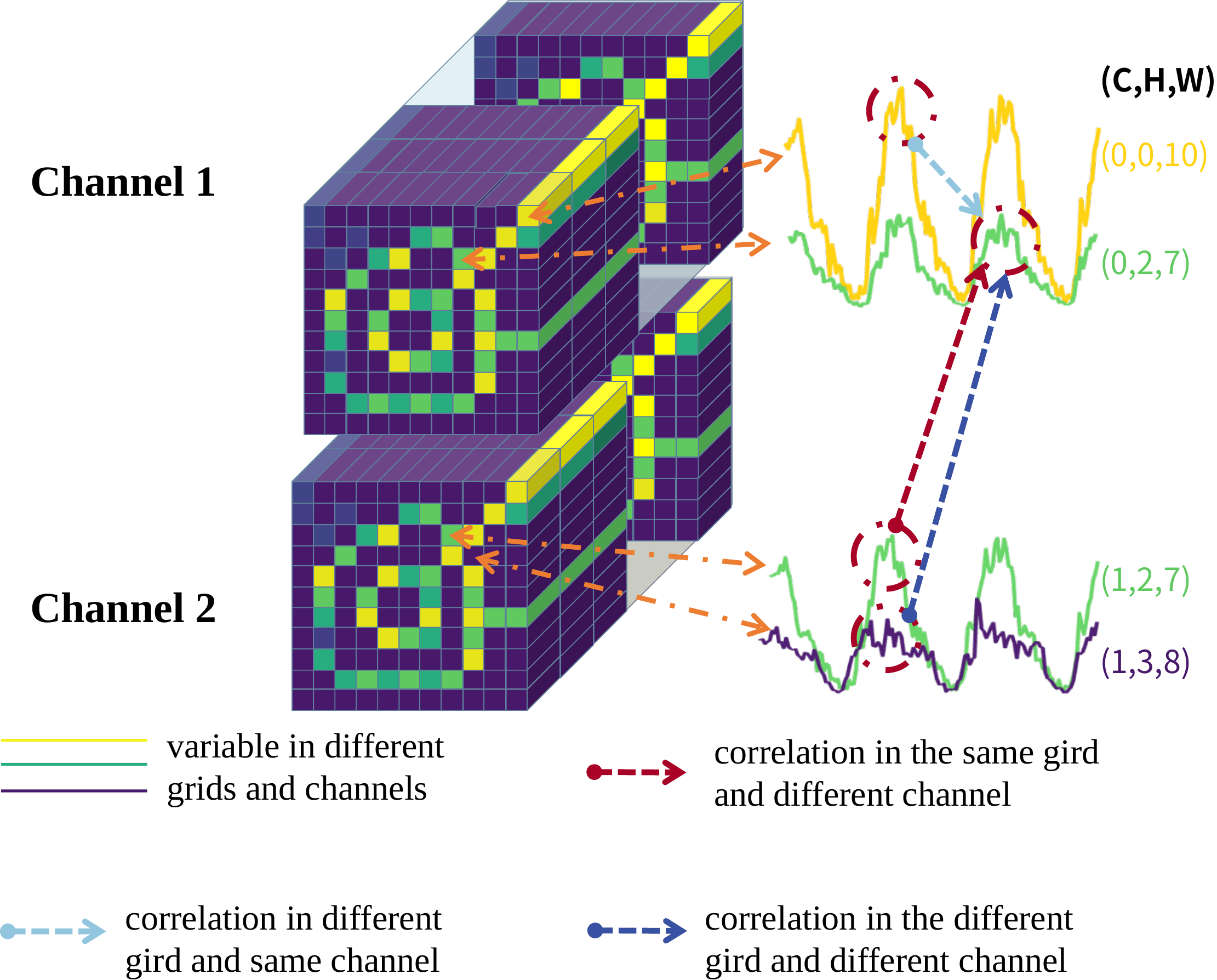} }
    \hspace{5pt}  \caption{Illustration depicting distinct perspectives on grid-based urban mobility data: (a) The image patch and/or convolution reception field mix both channels and variables. 
To simplify the illustration, we only presented 2D convolution and 2D patch partition; (b) Each variable within each spatial grid and across all channels is treated as an independent entity. We explore three distinct types of correlations to generate the embeddings.}
    \label{fig1}
    %图B的后面要列一些找到的参考文献
\end{figure*}

Both CNN and ViT-style video prediction frameworks aim to capture cross-channel and spatial correlations by treating small patches of the image as a unified entity. In CNNs, this is achieved through 2D/3D convolution filters~\cite{simvp,tau,convlstm,e3dlstm,crevnet}, while in ViTs,  the input data is divided into non-overlapping 2D/3D patches~\cite{vivit,videoswintransfomer,tang2023swinlstm}, which are then used as input tokens for the Transformer model. This approach is particularly apt when dealing with images and video data. In the realm of visual data, adjacent pixels' RGB values merge into a cohesive whole within specific regions. This holistic perspective facilitates the extraction of diverse features, objects, and semantic insights. Merging RGB channels with image regions enriches the analysis, empowering deep learning models to capture meaningful features from the visual data~\cite{ulyanov2018deep, d2021convit}. However, this approach may not be well-suited for urban mobility data where each channel (analogous to RGB channels) stores a specific attribute for a particular region, and the attributes of each region carry unique semantic meanings. Moreover, these attributes may exhibit complex spatiotemporal correlations that should not be hastily dismissed. Adopting conventional CNN and ViT methods to them without thoughtful consideration could potentially disrupt and neglect significant cross-channel and spatial correlations~\cite{revisitingCNN}.   %What we are getting at is that, within urban mobility data, each channel's time series under every pixel should be treated as an independent element, requiring a specific structure to capture their intricate correlations.

In this paper, \textbf{\textit{we contend that grid-based urban mobility data should be treated as a super-multivariate time series rather than as video data}}. ``{Super-multivariate time series}'' refers to a time series characterized by a substantial number of variables and long-term temporal observations. We compare the video and super-multivariate time series perspectives of grid-based urban mobility data in Fig.\ref{fig1}. In Fig.\ref{fig1b}, both CNNs and ViTs employ a shared approach, wherein they tend to mix the adjacent channels and grid variable. However, this perspective might overlook some crucial correlations. For example, consider a grid cell with many working offices surrounded by grids with numerous entertainment venues. The outflow from this working grid during the evening commute might be highly correlated with the inflow to a distant residential grid. Furthermore, the grids surrounding the corresponding residential grid may also exhibit significantly different characteristics compared to the residential grid. If we simply combine the inflow and outflow of these two grids and their surrounding grids into patches, it would likely lead to the oversight of the correlation related to the evening commute between the distant working and residential grids. In Fig.\ref{fig1a}, we present our perspective on urban mobility data. Instead of mixing channels and grid data into compact patches, we treat each channel's time series beneath every pixel as an independent entity. This approach allows us to directly utilize correlations within individual time series, both within the same grid and across different channels, as well as across different grids and channels, to generate embeddings. % 需要更详细的描述，比如说一个办公区域附近都是公园，它下班的outflow会和很远的某个居住小区的inflow高度相关，而这个inflow区域周边也都是属性很不相同的grid，如果将这两个小区和其周边区域混合在一起产生embedding，显然会使得下班回家这一correlation被忽略掉.

Additionally, our focus is on addressing the long-term prediction of urban mobility data, in contrast to the prevalent practice found in standard video prediction paradigms, where typically only a few steps ahead are predicted (e.g., 4 steps, as seen in most literature~\cite{simvp, tang2023swinlstm, guen2020disentangling, yu2019efficient}). Our emphasis on long-term prediction stems from its essential role in many urban management scenarios~\cite{informer}. Long-term predictions offer management personnel the lead time needed for effective preparation and planning. This extended forecasting horizon is critical for enabling proactive decision-making, ensuring that administrators are well-prepared to address future challenges and opportunities~\cite{zheng2014urban,boyce2015forecasting}. 

To this end, we introduce a novel \textbf{Super-Mutlviarate Urban Mobility Transformer (SUMformer)} for this task. We begin by converting the video data with dimensions $(T, C, H, W)$ into a super-multivariate time series with $C \times H \times W$ variables. Next, we aggregate the time steps along the temporal dimension, organizing them into subseries-level patches for subsequent temporal, variable, and frequency (TVF) blocks of SUMformer. In the temporal dimension, SUMformer offers two options for capturing the temporal relationships between patches: one relies on a pure MLP, while the other is based on self-attention. In the variable dimensions, given a multitude of time series in grid-based mobility data, our specific design incorporates an efficient self-attention mechanism inspired by \cite{set_transformer,linformer,additive_attention} to compute cross-variable attention. Notably, this design achieves a computational time complexity that scales linearly with respect to the number of variables involved. Lastly, in the frequency domain, recognizing that low-frequency periodic information holds more long-term predictive information, SUMformer employs Fourier low-frequency filters to process the features. We organize the hierarchical TVF blocks together using the patch merge approach of the Swin Transformer~\cite{swintransformer}, enabling SUMformer to capture multi-scale spatiotemporal and cross-variable correlations. By integrating these carefully-designed components, SUMformer achieves state-of-the-art (SOTA) performance in long-term urban mobility prediction.

Our contributions to the long-term urban mobility prediction challenge using SUMformer are as follows:
\begin{itemize}
\item We present a novel super-multivariate perspective on grid-based urban mobility data. Through this approach, we are able to utilize general multivariate time series forecasting models to achieve long-term urban mobility predictions.
\item We present the SUMformer: a Transformer model designed to leverage temporal, frequency, and cross-variable correlations for urban mobility forecasting. Notably, it stands out as one of the few Transformer models that explicitly taps into and harnesses cross-variable correlations across every channel and grid for urban mobility prediction.
\item Experiments (detailed in Section~\ref{sec:ex}) demonstrate that SUMformer surpasses state-of-the-art methods across three real-world datasets. We emphasize the significance of the super-multivariate perspective, explicit cross-variable correlation modeling, and frequency information for achieving optimal performance.
\end{itemize}

\section{Related Work}

\subsection{Urban Mobility Prediction as Video Prediction}
%从空间特征提取和时间特征提取两方面说
%空间特征提取，主流的就是CNN-based，Patch-based方法，我们认为它们都不好的原因是，直接使用video的那一套，忽视了交通流数据的特殊性，无论是CNN还是Patch-based，都会打破网格的独立性。
%时间特征提取，有两种不同的策略，RNN-based的自回归方式，One-Shot manner方式，我们认为RNN不适合长时预测，而One-shot那一套倒是有很好的性能。
%接下来说有很多视频预测模型直接应用到了交通流上。
Urban mobility prediction has garnered attention in machine learning recently. Initial research largely focused on CNN-based methodologies~\cite{lin2019deepstn+,predcnn,zhang2020understanding,wang2017predrnn,wang2018predrnn++,wang2019memory}, stemming from their success in image processing, as demonstrated by~\cite{zhang2017deep}. While models like SimVP~\cite{simvp} predominantly utilize CNNs, newer works argue against their efficiency in capturing global spatial dependencies~\cite{revisitingCNN}. Such concerns led researchers to introduce enhancements like the ConvPlus structure in DeepSTN+ to address CNN's limitations in handling long-range spatial dependencies~\cite{lin2019deepstn+}. Meanwhile, the emergence of the Transformer~\cite{vaswani2017attention} in NLP has influenced computer vision studies. ViT~\cite{ViT}, for instance, adapted Transformer techniques for visual tasks. Its patch-based processing approach inspired other models, such as the MLP-Mixer~\cite{mlpmixer}, which segments images into patches and processes them using a standalone MLP architecture. These patch-based strategies have also been adopted in spatial-temporal forecasting~\cite{tang2023swinlstm,mlpst}, including the application of large models like Pangu-Weather~\cite{pangu} in weather forecasting. Urban mobility datasets, notably TaxiBJ~\cite{zhang2017deep}, serve as common benchmarks for video prediction algorithms. Many frameworks are evaluated using these urban mobility datasets, including ~\cite{tau,simvp,mlpst,tang2023swinlstm}. However, these studies typically prioritize general video prediction, often focusing on short-term forecasts. This contrasts with the urban management requirement for predicting mobility trends days ahead.

%In some studies~~\cite{zonoozi2018periodic, yao2019revisiting}, Recurrent Neural Networks (RNNs) are employed to model periodic temporal dependencies. ST-GAN~~\cite{zhang2020understanding} employs a generative adversarial training strategy to train a CNN, enabling the learned model to generate realistic simulations. %按照这个逻辑再填写一些参考文献
%also 随着transformer在NLP领域取得了成功，一些研究发现将图片切为patch作为token输入到transformer结构中在图片分类，物体识别，分割等等下游任务中取得了非常好的效果。同时又有一些研究提出，将patch机制输入到纯MLP的结构依然能取得强劲的性能。

%把你找到的文献填进去.

% 需要再说明一下，现有的视频预测方法破坏了独立性

%Representation learning, understanding, and prediction of videos, especially natural videos, have garnered significant attention. Due to the intricate spatiotemporal correlations in video data, a substantial amount of research focuses on designing network structures to enhance the network's ability to extract features from videos, employing architectures such as RNNs~\cite{e3dlstm,crevnet,phydnet,convlstm}, CNNs~\cite{simvp,tau,predcnn,Photo-realistic}, and Transformers~\cite{viViT,MViT,videoswintransfomer}. 这句话没有体现CNN经常被使用 For prediction tasks, one-shot methods, compared to autoregressive methods, are simpler in both network structure and training processes. Some competitive methods use pure CNN structures to spatially downsample input video data and perform 2D convolutions for extracting object features in the video. In the temporal dimension, they capture changes of objects using 1D convoluions and output prediction in a one-shot manner.

\subsection{Multivariate time series forecasting Framework}

Deep neural networks (DNNs), particularly Transformer models, have significantly advanced time series forecasting, emphasizing long-term predictions since pioneering works like Informer~\cite{informer}. Multivariate time series forecasting's success hinges on modeling cross-variable correlations. Broadly, methods are classified into variable-dependent strategies~\cite{informer,autoformer,fedformer,sageformer} and variable-independent strategies~\cite{TCN,Nlinear,patchtst}. To clarify terminology, we use ``variable'' instead of ``channels'' as in~\cite{patchtst}. {Variable-dependent methods treat time series comprehensively, with the majority rudimentarily mapping the cross-variable dimension at the same time step to a latent space for \textbf{implicit} modeling.} Yet, they have been critiqued for inconsistency during distribution shifts among variables~\cite{Channel-Indepence}. On the other hand, variable-independent methods~\cite{Nlinear,patchtst} apply univariate models across multiple correlated variables. Despite neglecting correlations, they have shown enhanced performance~\cite{Channel-Indepence}. However, this strategy can yield suboptimal forecasts due to limited capacity~\cite{Channel-Indepence}. A special method is Crossformer~\cite{crossformer}, which leverages self-attention mechanism to \textbf{explicitly} explore cross-variable correlations, achieving good performance in general time series forecasting task. In urban mobility data, high-resolution grids produce a multitude of time series. Areas with similar semantic and geographical features tend to have strong correlations. Moreover, the high granularity of grids can result in a super-multivariate time series. This complexity makes capturing correlations with computational-heavy variable-dependent strategies a challenge. Efficient attention mechanisms, such as those found in ~\cite{set_transformer,linformer,additive_attention}, are crucial to address this issue.

\subsection{Deep learning model leveraging frequency-domain information}

% 先参考chunwei的那个arxiv论文，把其它模型怎么用傅里叶信息加上，然后再稍微写一些傅里叶算子具备很强的能力
The frequency domain analysis algorithm like Fast Fourier Transform (FFT) converts data from the time domain to the frequency domain and serves as a frequency-domain feature extraction module in constructing neural network architectures~\cite{fedformer,timesnet,koopa}. Initially proposed as a data-driven method for solving partial differential equations (PDEs), the Fourier Neural Operator (FNO)~\cite{FNO} has subsequently proven effective in image classification~\cite{FNOimage} and time series forcasting~\cite{autoformer,fedformer,timesnet,yang2023enhancing}. 
FourcastNet~\cite{pathak2022fourcastnet} accurately captures the formation and movement of weather patterns through the utilization of the adaptive Fourier neural operator (AFNO). CoST~\cite{woo2022cost} leverage contrastive learning methods, transform the input series into frequency domain to learn discriminative seasonal representations.
TimesNet~\cite{timesnet} transforms the time series into 2D space based on multiple periods and applied a 2D kernel for features extraction. Even though urban mobility data clearly shows strong periodic patterns\textemdash a significant frequency domain feature\textemdash few studies have tapped into this frequency domain information for urban mobility prediction.

%加一点CoST等无监督频率方式
%Given the pronounced temporal periodicity in traffic flow, we opt to evaluate FNO1d~\cite{FNO} and FNO3d~\cite{FNO} to model traffic flow as videos and super-multivariate time series respectively in the subsequent experimental section, assessing their feature extraction and prediction performance. Our proposed model draws inspiration from FNO's approach of sampling the lower frequencies in the spectrum modes. We devised a low-frequency filtering module as a successor to the univariate temporal module and cross-grids attention module. This design keeps meaningful information features from lower Fourier modes and filters out the higher modes. % incorporate 已有的方法显得太low了，你可以把自己的算子叫做低频滤波器 we design a low-frequency filtering 之类的
%While FNO focus on applying linear transform on lower Fourier modes and filters out higher modes, we designed a low-frequency filter module which 
%, we devised a low-frequency filtering module as a successor to the univariate temporal module and cross-grids attention module. This design keeps meaningful information features from lower Fourier modes and filters out the higher modes.

\section{Methodology}
% \begin{figure*}[t]
%   \centering
%   \includesvg[width=0.9\textwidth]{Arch&Blocks}
%   \caption{Architecture.}
%   \label{fig1}
% \end{figure*}
\begin{figure*}[t]
\centering    
\subfigure[Data Flow]{
\label{fig2a}
\includegraphics[width=0.93\linewidth]{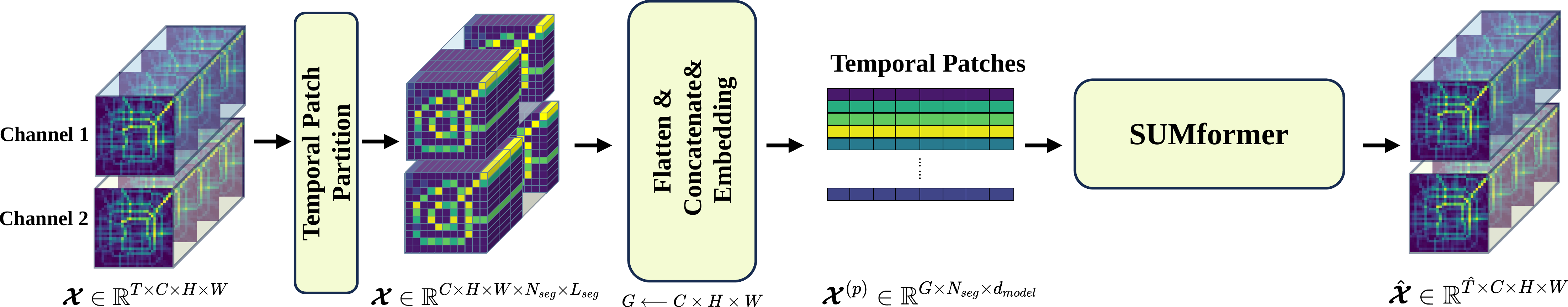}}
\centering    
\subfigure[Architecture]{
\label{fig2b}
\includegraphics[width=0.69\linewidth]{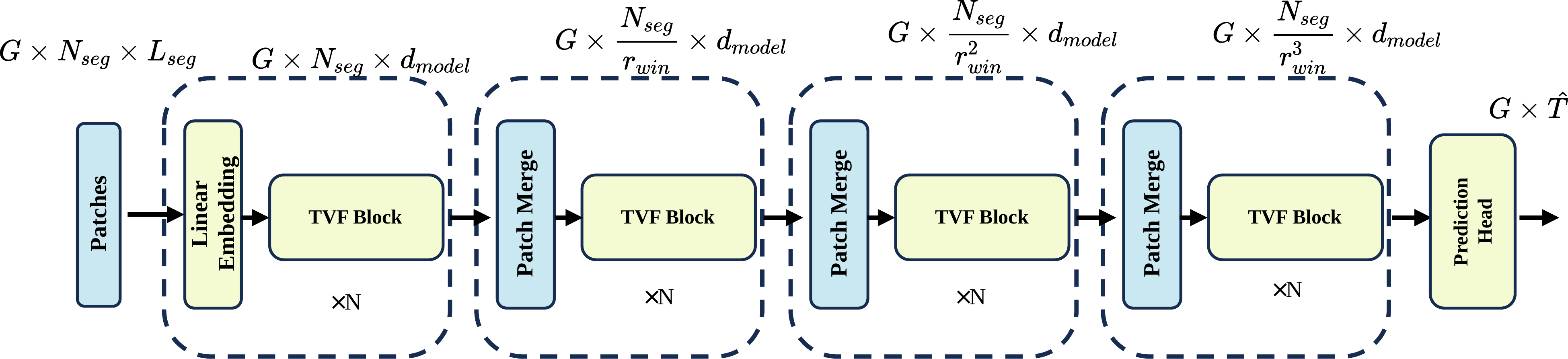}}
\qquad
\subfigure[A single TVF block]{
\label{fig2c}
\includegraphics[width=0.24\linewidth]{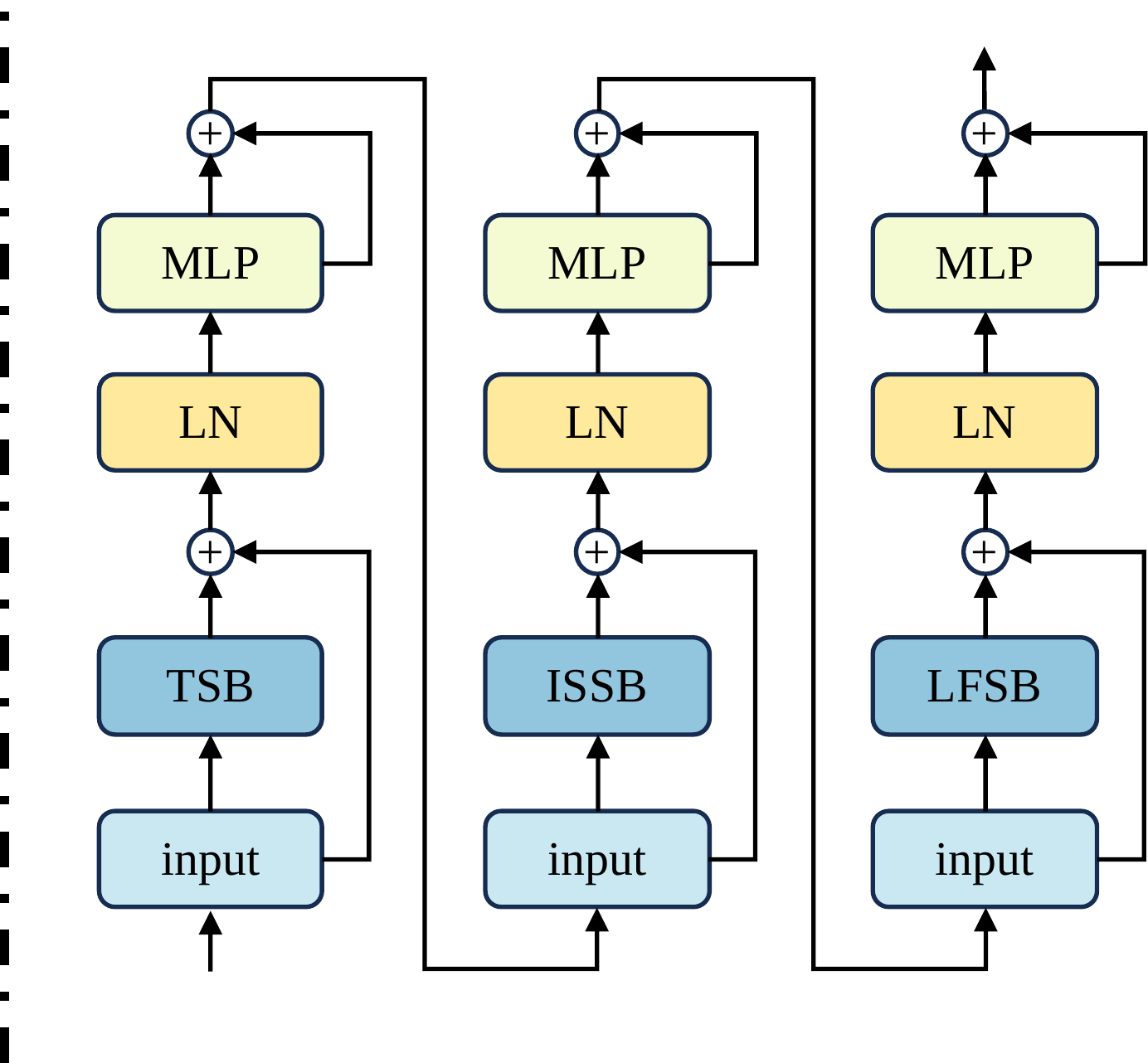}}
\caption{(a) Data flow in a SUMformer, where grid-based mobility data is flattened into super-multivariate patches before processing; (b) Architecture inspired by the Swin Transformer~\cite{swintransformer}; (c) A single TVF (Temporal-Variate-Frequency) block, consisting of a temporal sub-block, an Inter-Series Sub-Block, and a Low-frequency Filter Sub-Block.}
\label{fig2}
\end{figure*}

\subsection{Overall Architecture}
%讲编码方式，讲patch机制，再讲我用了三个模块分别行使不同功能，最后讲patch merge提升长时性能。
%The overall workflow of our method is illustrated in Fig~\ref{fig2}.
Fig.~\ref{fig2} presents an overview of the SUMformer architecture. The core elements of SUMformer encompass a temporal patch mechanism for generating super-multivariate temporal patches from input videos, the TVF block, which fully exploits temporal, cross-variable (inter-series), and frequency-domain information, and the temporal patch merging mechanism for capturing multi-scale correlations. In the following sections, we will provide a detailed introduction to all the components of SUMformer.

\subsubsection{\textbf{Temporal Patch Partition}} {In the temporal dimension, we begin by segmenting the input into patches at the sub-series level, with each patch functioning as a token input for the SUMformer.} The input is denoted as $\boldsymbol{\mathcal{X}}\in\mathbb{R}^{T\times C\times H\times W}$, where $T$ is the number of time steps, $C$ is the number of variables per frame, and $H$, $W$ are the height and width of the frame. First, we flatten it into a super-multivariate time series denoted as a $\boldsymbol{X}\in\mathbb{R}^{G\times T}$, where $G = C\times H\times W$. Subsequently, it is sliced into non-overlapping sub-series $\boldsymbol{\mathcal{X}}'\in\mathbb{R}^{G\times N_{seg}\times L_{seg}}$, where $N_{seg}$ is the number of sub-series $N_{seg} = \frac{T}{L_{seg}}$. Through a linear embedding layer shared by $G$ variables and $N_{seg}$ segments, the length of each patch is projected to a fixed dimension of $d_{model}$:
\begin{equation}
    \boldsymbol{x}_{i,j}^{(p)} = \boldsymbol{x}'_{i,j}\boldsymbol{W}_{patch}+\boldsymbol{W}_{pos}, 1\le i\le G, 1\le j\le N_{seg},
    \label{patch-transfer}
\end{equation}
where $\boldsymbol{x}_{i,j}^{(p)}\in\mathbb{R}^{d_{model}}$ denotes the embedding transferred from the original $j$-th patch of the $i$-th sub-series; $\boldsymbol{W}_{patch}$ denotes a linear projection layer; $\boldsymbol{W}_{pos}\in\mathbb{R}^{G\times N_{seg}\times d_{model}}$ denotes a learnable parameter as the positional embedding. Following the linear embedding layer, we obtain a tensor $\boldsymbol{\mathcal{X}}^{(p)}\in\mathbb{R}^{G\times N_{seg}\times d_{model}}$. In contrast to CNN or ViT-based methods, we do not downsample or divide it into spatial patches. Instead, we treat it as a super-multivariate time series to maintain the independence of information between grids and channels. We verify the advantages of preserving this grid independence in our subsequent experiments.

\subsubsection{\textbf{TVF Block}} 
The input tensor $\boldsymbol{\mathcal{X}}^{(p)}$ then passes through a series of stacked Temporal-Variable-Frequency (TVF) blocks, as depicted in Fig.~\ref{fig2c}. Each of these blocks comprises three sub-blocks designed for feature processing from the temporal, inter-series, and frequency domains, respectively. Each sub-block comprises a corresponding processing module, followed by a LayerNorm (LN) layer and a 2-layer MLP using GELU activation. Residual connections are applied for each layer:
\begin{equation}
	\begin{split}
	&\hat{\boldsymbol{\mathcal{X}}}^{(p)l} = \text{LayerNorm}\left(\boldsymbol{\mathcal{X}}^{(p)l}+sub\text{-}block\left(\boldsymbol{\mathcal{X}}^{(p)l}\right)\right),\\
	&\boldsymbol{\mathcal{X}}^{(p)l+1} = \text{LayerNorm}\left(\hat{\boldsymbol{\mathcal{X}}}^{(p)l}+ MLP\left(\hat{\boldsymbol{\mathcal{X}}}^{(p)l}\right)\right),
	\end{split}
 \label{sub-block}
\end{equation}
where $\boldsymbol{\mathcal{X}}^{(p)l+1}$ and $\boldsymbol{\mathcal{X}}^{(p)l}$ represent the input and output of the $l$-th layer,  respectively.

\subsubsection{\textbf{Patch Merging}} To capture long-term temporal correlations within the super-multivariate time series, the number of patches is reduced through the patch merging~\cite{swintransformer} mechanism, as depicted in Fig.~\ref{fig2b}. The patch merging layer concatenates the features of each group of adjacent fixed-size windows and applies a linear layer to the $N_{seg}$-dimensional concatenated features. With each merging operation, The patch size of the input tensor is reduced by a factor of $r_{win}$, which denotes the number of merged sub-series. With several layer of patch merging, the model's temporal receptive field grows exponentially which is advantageous for effectively capturing correlations at different scales. Eventually, all sub-series in the same variable are fused into a single token. After passing through a linear prediction layer, we yield the prediction results.
\subsection{Temporal Sub-Block}
%分开讲不同模块，其中二最多，三其次，一最少。
%一就按照MSA那一套去讲，或者我还可以说说我用MLP-mixer做时序头，放到论文里，辅助证明我和Crossformer patchTST那一套是不一样的。嗯 后面可以做实验验证一下。
The main objective of this module is to capture temporal correlations within each individual univariate time series, with all time series sharing the same set of parameters. To achieve this, we have introduced two options, one based on Multi-Head Self Attention (MHSA)~\cite{vaswani2017attention} and the other on MLP-Mixer~\cite{mlpmixer}.

\subsubsection{\textbf{TSB-MHSA}} 
In this version, we utilize MHSA to extract the correlation among sub-series within the same variable. The patches from all time series, denoted as $\boldsymbol{\mathcal{X}}_{i,:}^{(p)}\in\mathbb{R}^{N_{seg}\times d_{model}}$, are initially mapped by $\boldsymbol{W}_Q^{(h)}$, $\boldsymbol{W}_K^{(h)}$, and $\boldsymbol{W}_V^{(h)}$ into $\boldsymbol{Q}_{i,:}^{(h)}$, $\boldsymbol{K}_{i,:}^{(h)}$, and $\boldsymbol{V}_{i,:}^{(h)}\in\mathbb{R}^{N_{seg}\times d_{qkv}}$ within the latent space, respectively. Here, $1\le i\le G$, and $h$ represents the current head number of the MHSA. Then, we have
\begin{equation}
        {\boldsymbol{{Z}}_{i,:}^{(h)}}
        = \text{Softmax}\left(\frac{\boldsymbol{Q}_{i,:}^{(h)}{\boldsymbol{K}_{i,:}^{(h)}}^{T}}{\sqrt{d_{qkv}}}\right)\boldsymbol{V}_{i,:}^{(h)},
    \label{MHSA}
\end{equation}
where $d_{qkv}$ is a constant, the $h$ heads, denoted as ${\boldsymbol{{Z}}_{i,:}^{(h)}}$, together form a tensor $\boldsymbol{\mathcal{Z}}_{i,:}\in\mathbb{R}^{N_{seg}\times h\times d_{qkv}}$. Subsequently, the output, $\boldsymbol{\mathcal{Z}}_{i,:}\in\mathbb{R}^{N_{seg}\times h\times d_{qkv}}$, is flattened along the dimension of the head number and then mapped through linear output projection to produce the final output $\boldsymbol{{O}}^{(time)}_{i,:}\in\mathbb{R}^{N_{seg}\times d_{model}}$.

\subsubsection{\textbf{TSB-MLP}} We also offer a pure-MLP architecture as an alternative to MHSA. We utilize two two-layer MLP networks, each incorporating the GELU activation function, Dropout, and residual connections to capture the internal features of the sub-series and the features spanning across sub-series. When provided with the input $\boldsymbol{\mathcal{X}}_{i,:}^{(p)}$, the process is as follows:
\begin{equation}
    \begin{split}
        &\hat{\boldsymbol{\mathcal{Z}}}_{i,j,:} = \boldsymbol{\mathcal{X}}_{i,j,:}^{(p)} + \boldsymbol{W}_2\sigma\left(\boldsymbol{W}_1\text {LayerNorm}\left(\boldsymbol{\mathcal{X}}_{i,j,:}^{(p)l}\right)\right),\\
        &\boldsymbol{{O}}^{(time)}_{i,:,k} = \hat{\boldsymbol{{Z}}}_{i,:,k}+\boldsymbol{W}_4\sigma\left(\boldsymbol{W}_3\text {LayerNorm}\left(\hat{\boldsymbol{\mathcal{Z}}}_{i,:,k}\right)\right),
    \end{split}
    \label{MLP-mixer}
\end{equation}
where $1\le i\le G$, $1\le j\le N_{seg}$ and $1\le k\le d_{model}$; $W_n(n=1,2,3,4)$ denote the learnable weight matrices, $\boldsymbol{{O}}^{(time)}\in\mathbb{R}^{G\times N_{seg}\times d_{model}}$ denotes the output of TSB-MLP sub-block.

\subsection{Inter-Series Sub-Block }
\begin{figure}[t]
\centering    
\subfigure[Full attention]{
\label{fig3a}
\includegraphics[width=0.45\linewidth]{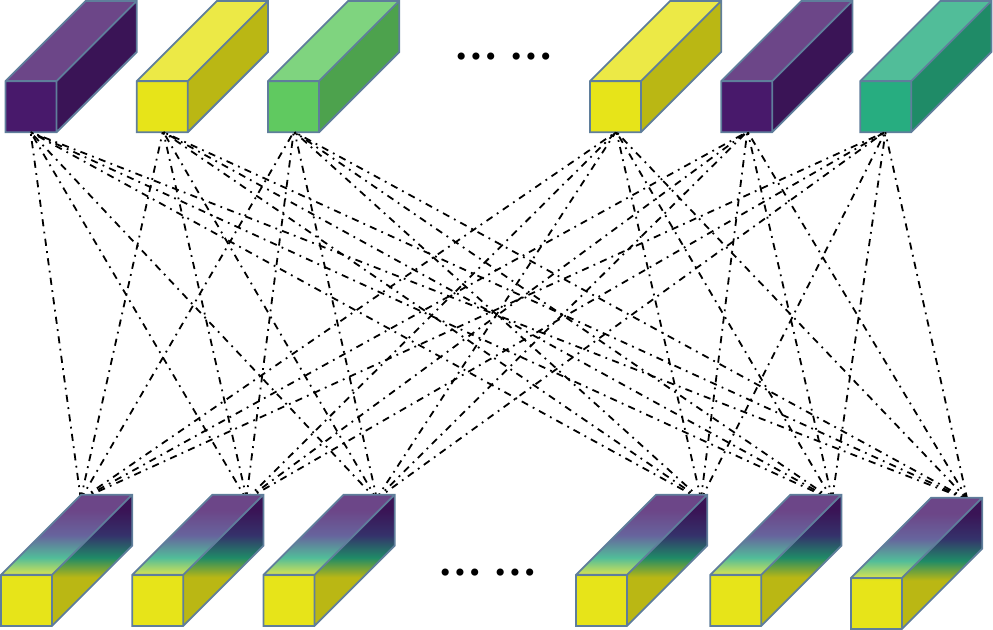}}
\subfigure[Neural dictionary]{
\label{fig3b}
\includegraphics[width=0.5\linewidth]{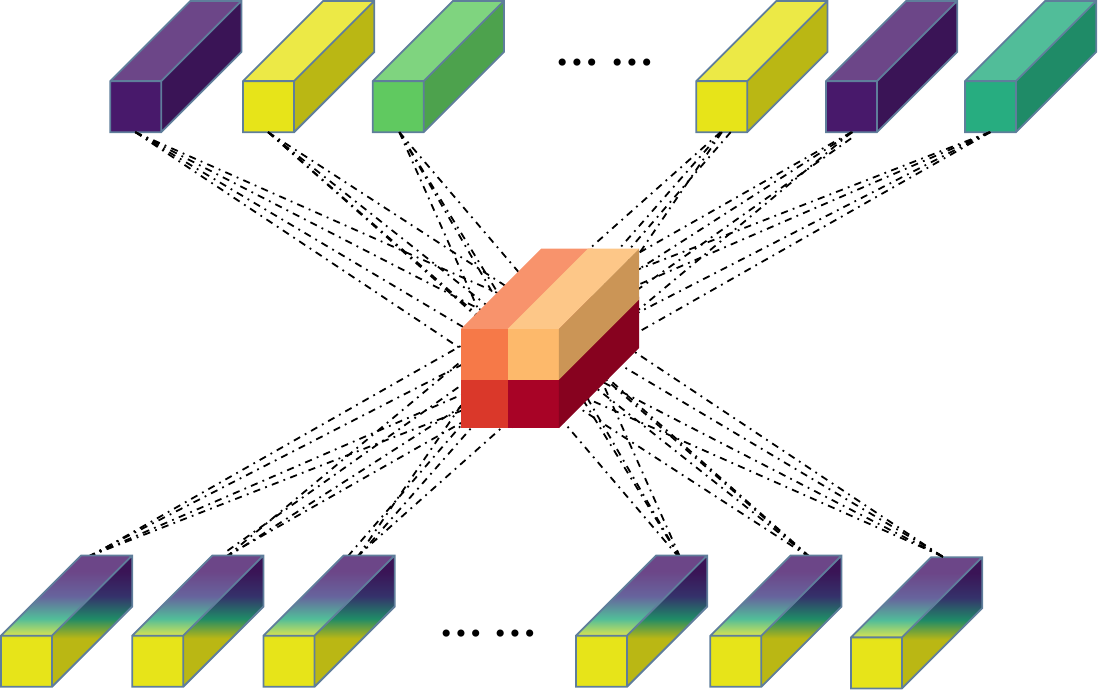}}
\subfigure[Low-rank projection]{
\label{fig3c}
\includegraphics[width=0.45\linewidth]{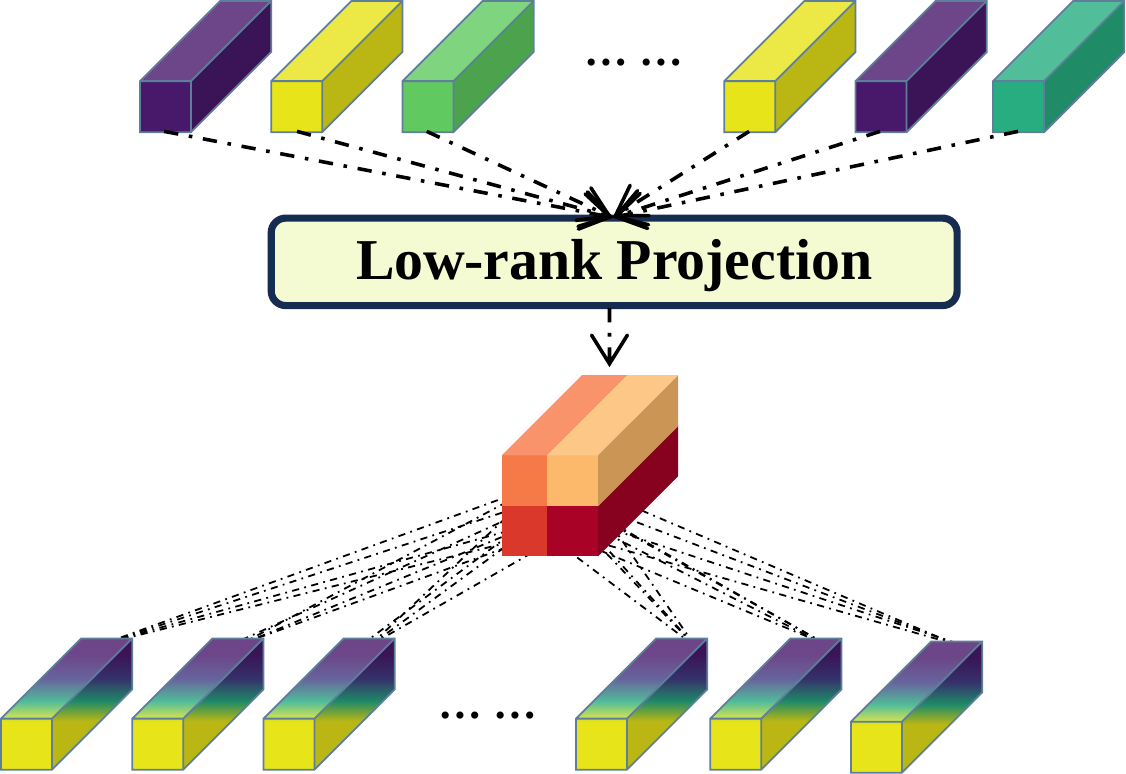}}
\subfigure[Additive attention]{
\label{fig3d}
\includegraphics[width=0.5\linewidth]{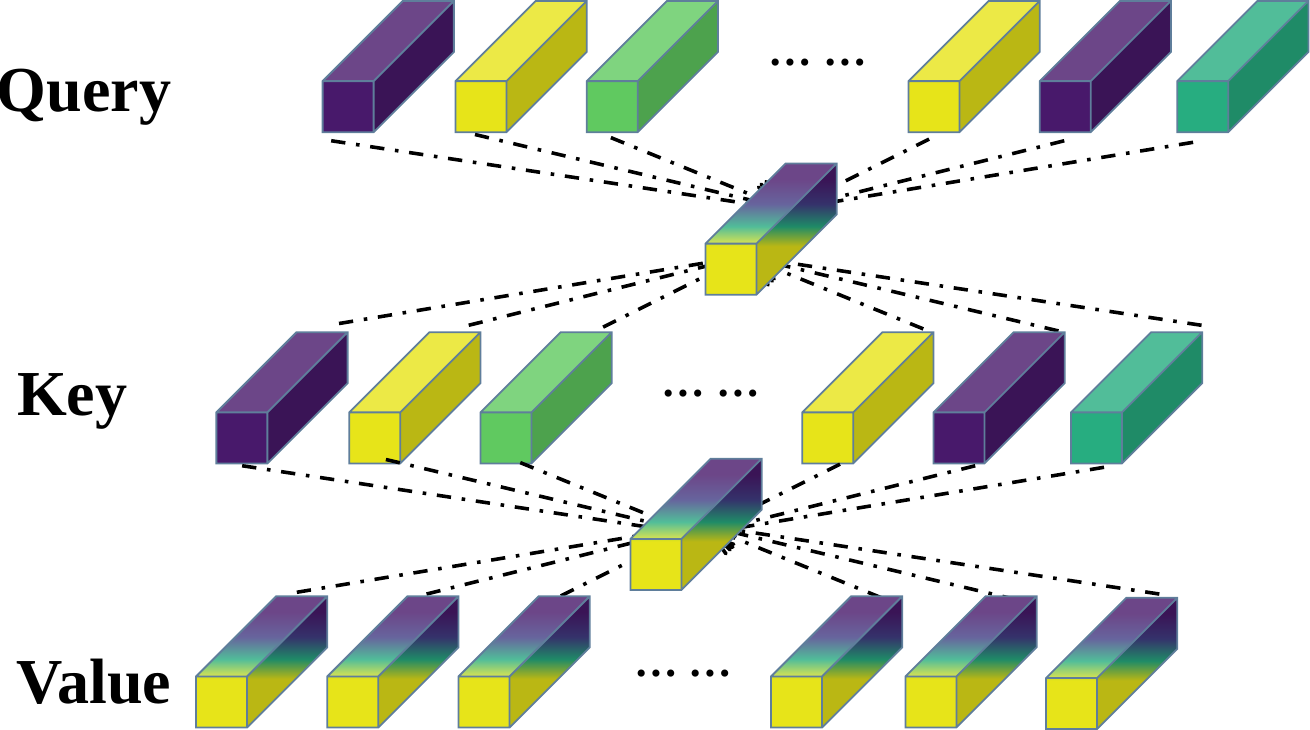}}
\caption{Alternative Attention Mechanism Choices for the Inter-Series Sub-Block.}
\label{fig3}
\end{figure}

%二就只能是使用router那一套来讲，然后其实可以阐述一下这种方式，对于不同grids，不同variable的影响，可能用到数学。
%介绍三种linear attention在我们模块里的应用low-rank projection attention, memory-based attention, additive attention三个版本
%A multitude of correlations exists among different mobility variables, encompassing inter-regional correlations, intra-regional variable correlations, and inter-regional variable correlations. By flattening both the channel and spatial dimensions, the model can efficiently learn to capture these three types of correlations with a well-designed approach. To address this, we introduce an Inter-series Sub Block (ISSB). It's worth noting that in super-multivariate time series data, like TaxiBJ, with the number of series after unfolding denoted as $G=2048$, direct computation of MSA results in a high computational complexity of $O(G^{2})$, as shown in \ref{fig4a}. 
The purpose of the Inter-Series Sub-Block (ISSB) is to capture correlations between different variables. In the case of urban mobility data, a significant challenge arises due to the potentially large number of time series present in an urban mobility video. For instance, the popular benchmark dataset TaxiBJ comprises a total of 2048 time series. It's important to note that this number can be even larger in real-world scenarios. For example, we may use finer latitude and longitude resolutions to define more detailed fine-grained videos~\cite{liang2021fine} (larger $H$ and $W$), or incorporate additional mobility attributes for each grid, such as the inflow and outflow of various travel modes (larger $C$). We can certainly use the raw transformer proposed in~~\cite{vaswani2017attention} to capture correlations between variables (Fig.~\ref{fig3a}), but its computational cost is $O(G^2)$, which would make the model computationally burdensome for ``hyperspectral fine-grained'' urban mobility video data. Instead, SUMformer offers several alternative options with $O(G)$ complexity, as illustrated in Fig.~\ref{fig3}. 
%In order to reduce complexity and inspired by Set transformer\cite{set_transformer}, Linformer\cite{linformer} and Fastformer\cite{additive_attention}, we designed three efficient attention modules, lowering the computational complexity to $O(G)$.
%{\color{red}To reduce computational complexity, we employ an asymmetric bidirectional self-attention mechanism (ABSA). Instead of directly computing MSA, we utilize a set of learnable parameters, referred to as the bottleneck messenger, to traverse all patches with a fixed dimensionality denoted as $d << G$. Using this mechanism, the final computational complexity is reduced to $O(G)$.}

\subsubsection{\textbf{Neural Dictionary}}
The neural dictionary model, proposed in~\cite{set_transformer}, utilizes a learnable neural dictionary $\boldsymbol{Dic}\in\mathbb{R}^{g\times d_{model}}$ with a fixed number $g$ (where $g \ll G$) of vectors to reduce computational complexity. For an input $\boldsymbol{\mathcal{X}}_{:,i}^{(p)l}\in\mathbb{R}^{G\times d_{model}}$ from the $i$-th patch, we have
\begin{equation}
    \begin{split}
        &\boldsymbol{M} = \text{MHSA}\left(\boldsymbol{Dic},\boldsymbol{\mathcal{X}}_{:,i}^{(p)l},\boldsymbol{\mathcal{X}}_{:,i}^{(p)l}\right),\\
        &\boldsymbol{{O}}^{(spatial)}_{:,i} = \text{MHSA}\left(\boldsymbol{\mathcal{X}}_{:,i}^{(p)l},\boldsymbol{M},\boldsymbol{M}\right),
    \end{split}
    \label{neural-dictionary-eq}
\end{equation}
where we employ the neural dictionary $\boldsymbol{Dic}$ as the query, and the input $\boldsymbol{{X}}_{:,i}^{(p)l}$ as the key-value pair in the first MHSA layer, creating an aggregated message denoted as $\boldsymbol{M}\in\mathbb{R}^{g\times d_{model}}$ with the same shape as $\boldsymbol{Dic}$. Subsequently, this $\boldsymbol{M}\in\mathbb{R}^{g\times d_{model}}$ is utilized as the key-value pair in the second MHSA layer to interact with the query $\boldsymbol{{X}}_{:,i}^{(p)l}$, resulting in the final output denoted as $\boldsymbol{{O}}^{(spatial)}_{:,i}\in\mathbb{R}^{G\times d_{model}}$.

\subsubsection{\textbf{Low-rank Projection}}
Unlike aggregating information using a neural dictionary and computing the correlations with the input $\boldsymbol{\mathcal{X}}_{:,i}^{(p)l}$ through MHSA, in the low-rank projection version~\cite{linformer}, we use a $(g\times G)$-dimensional projection matrix $\boldsymbol{W}_{lin}$ to map the key and value from $G$-dimension to $g$-dimension where $g$ is a fixed number and $g\ll G$ (see Fig.~\ref{fig3c}).
\begin{equation}
\begin{split}
    &\boldsymbol{\mathcal{\hat{X}}}_{:,i}^{(p)l} = \boldsymbol{W}_{lin}\boldsymbol{\mathcal{X}}_{:,i}^{(p)l},\\
    &\boldsymbol{{O}}^{(spatial)}_{:,i} = \text{MHSA}\left(\boldsymbol{\mathcal{X}}_{:,i}^{(p)l},\boldsymbol{\mathcal{\hat{X}}}_{:,i}^{(p)l},\boldsymbol{\mathcal{\hat{X}}}_{:,i}^{(p)l}\right),
\end{split}
\label{low-rank-eq}
\end{equation}
where all variable information is aggregated to a smaller size output denoted as $\boldsymbol{\mathcal{\hat{X}}}_{:,i}^{(p)l}\in\mathbb{R}^{g\times d_{model}}$ through $\boldsymbol{W}_{lin}$. Subsequently, we employ MHSA to calculate the correlation among the $G$ variables.

\subsubsection{\textbf{Additive Attention}}
The patches from all variables at the same time steps denoted as $\boldsymbol{\mathcal{X}}_{:,i}^{(p)}\in\mathbb{R}^{N_{seg}\times d_{model}}$, are initially mapped by $\boldsymbol{W}_Q^{(h)}$, $\boldsymbol{W}_K^{(h)}$, and $\boldsymbol{W}_V^{(h)}$ into $\boldsymbol{Q}_{:,i}^{(h)}$, $\boldsymbol{K}_{:,i}^{(h)}$, and $\boldsymbol{V}_{:,i}^{(h)}\in\mathbb{R}^{G\times d_{qkv}}$ within the latent space, respectively.
As shown in Fig.~\ref{fig3d}, the Additive Attention mechanism~\cite{additive_attention} first summarizes the query $\boldsymbol{Q}_{:,i}^{(h)}\in\mathbb{R}^{G\times d_{qkv}}$ with a 1-D mapping vector $\boldsymbol{w}_q\in\mathbb{R}^{d_{qkv}}$ using Softmax activation:
\begin{equation}
   \boldsymbol{\alpha} = \text{Softmax}\left(\frac{\boldsymbol{Q}_{:,i}^{(h)}\boldsymbol{w}_q}{\sqrt{d_{qkv}}}\right),
\end{equation}
where $\boldsymbol{\alpha}\in\mathbb{R}^{G}$ denotes the output query attention score. Then we get the global query vector $\boldsymbol{q}\in\mathbb{R}^{d_{qkv}}$ via $\boldsymbol{q} = \boldsymbol{\alpha}\boldsymbol{Q}_{:,i}^{(h)}$. Followed by an element-wise product between global query vector and key $\boldsymbol{K}_{:,i}^{(h)}\in\mathbb{R}^{G\times d_{qkv}}$, we model the correlation between query and key via $\boldsymbol{P}_{:,i}^{(h)} = \boldsymbol{q}*\boldsymbol{K}_{:,i}^{(h)}$, where $\boldsymbol{K}_{:,i}^{(h)}\in\mathbb{R}^{G\times d_{qkv}}$ and $*$ is the element-wise production. Subsequently, a similar procedure is employed to obtain a global key vector $\boldsymbol{k}\in\mathbb{R}^{d_{qkv}}$:
\begin{equation}
    \begin{split}
        &\boldsymbol{\beta} = \text{Softmax}\left(\frac{\boldsymbol{P}_{:,i}^{(h)}\boldsymbol{w}_k}{\sqrt{d_{qkv}}}\right),\\
        &\boldsymbol{k} = \boldsymbol{\beta}\boldsymbol{P}_{:,i}^{(h)}.\\
    \end{split}
\end{equation}
Then, an element-wise product is applied between value $\boldsymbol{V}_{:,i}^{(h)}\in\mathbb{R}^{G\times d_{qkv}}$ and global key vector to compute the key-value interaction $\boldsymbol{U}_{:,i}^{(h)}\in\mathbb{R}^{G\times d_{qkv}}$. Then we applied linear matrices and residual connection to obtain the final output of the inter-series correlation output $\boldsymbol{{O}}^{(spatial)}$:
\begin{equation}
\begin{split}
    &\boldsymbol{U}_{:,i}^{(h)} = \boldsymbol{k}*\boldsymbol{V}_{:,i}^{(h)},\\
    &\boldsymbol{{O}}^{(spatial)}_{:,i} = \boldsymbol{W}_1\left(\boldsymbol{Q}_{:,i}^{(h)}+\boldsymbol{W}_2\boldsymbol{U}_{:,i}^{(h)}\right),
\end{split}
\end{equation}
where $\boldsymbol{W}_1$ and $\boldsymbol{W}_2$ are learnable weight matrices.

%Furthermore, the inter-series correlation evolves with different time scales. In our approach, we initiate each patch with a relatively short temporal length, and this length exponentially increases with the patch merge mechanism. As a result, ISSB can effectively capture intercorrelation information across various scales. %这个可以可视化一下

%ABSA需要展开讲讲
%\begin{figure}[t]
%\centering    
%\subfigure[Full attention score]{
%\label{full_attention_score}
%\includesvg[width=0.45\linewidth]{full_attention_score}}
%\\
%\subfigure[Neural dictionary score]{
%\label{neuraldic_score}
%\includesvg[width=0.4\linewidth]{neuraldic_score}}
%\subfigure[Linear attention score]{
%\label{lin_score}
%\includesvg[width=0.4\linewidth]{lin_score}}
%\caption{XXXXXXXXXX}
%\label{attention_score}
%\end{figure} % 可视化分析应该放在最后，A点和B点之间的attention score很高，与B点周围的却不高，突出这个问题！，证明super-multivariate view的正确性，结合A点，B点及其周边区域的地理属性做分析

\subsection{Low-frequency Filter Sub-Block}
%三是着重讲一下我将patch 提前predict一下，转时域，low-frq-filter，再patch化，转patch域，交给下一组模块处理。实验证明提升了性能。LFT的公式部分可以做一下这个过程的展示

We consider periodicity within urban mobility data to play a substantial role in long-term prediction task. In frequency-domain analysis, it's well-established that periodicity primarily resides within the low-frequency components, while the high-frequency components introduce some level of noise. To effectively address this and bolster the model's resilience, we have introduced a Low-Frequency Filter Sub-Block (LFSB). This module serves the purpose of filtering out superfluous high-frequency elements, enhancing the model's robustness and preserving the essential periodic features.

\begin{figure}[htp]
\centering    
{\includegraphics[width=0.3\textwidth]{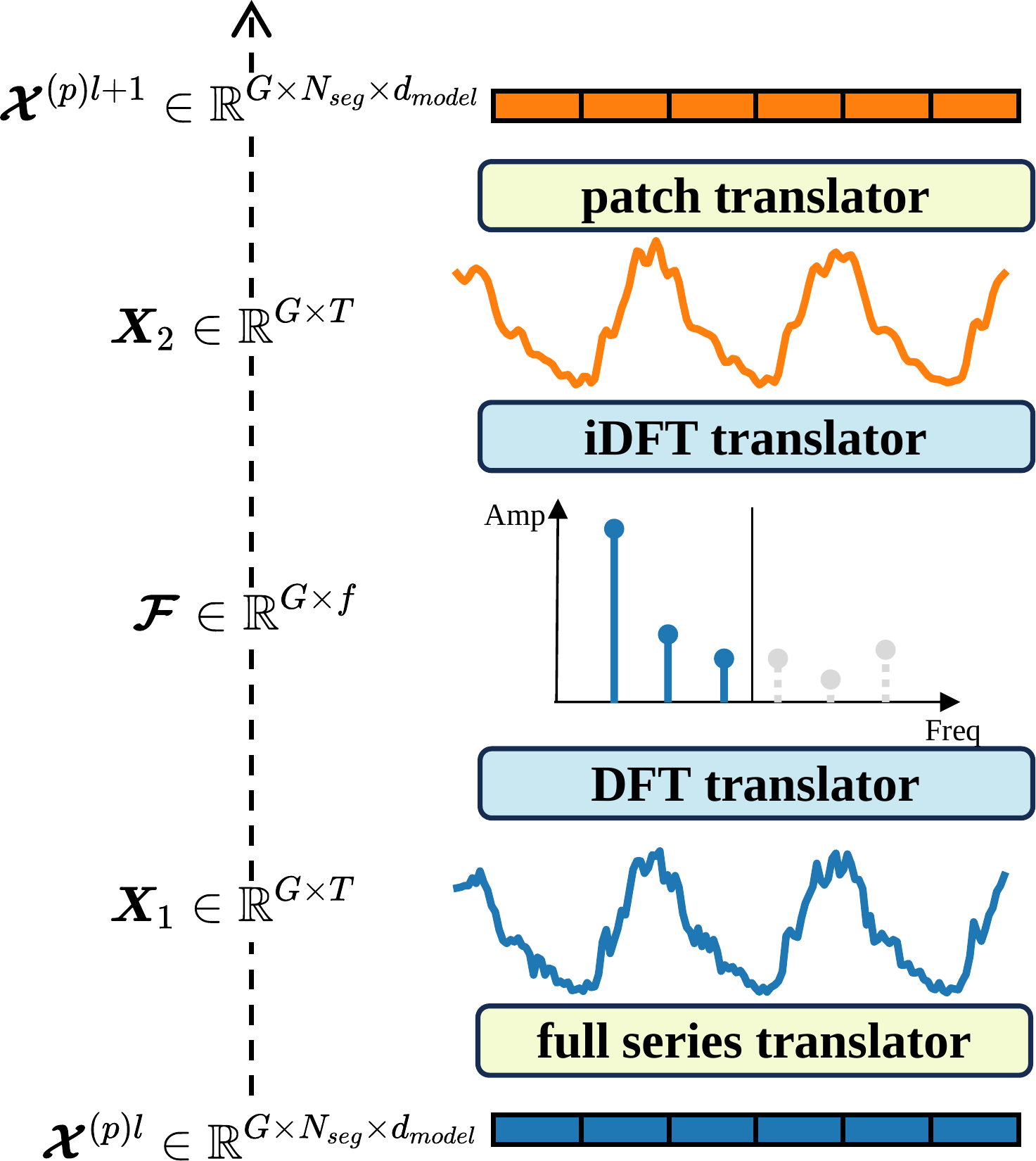}}
\caption{Illustration of Low-frequency Filter Sub-Block}
\label{LFSB}
\end{figure}

LFSB is composed of four translators, as illustrated in Fig.~\ref{LFSB}. Following TSB and ISSB, LFSB conducts low-frequency filtering within the frequency domain of the features generated by ISSB. The details are as follows:
\begin{equation}
  \begin{split}
    & \boldsymbol{X}_{1} = \boldsymbol{W}_{1} \text{Reshape}\left (\boldsymbol{\mathcal{X}}^{(p)l}\right), \\
    & \boldsymbol{\mathcal{F}} = \text{DFT}\left(\boldsymbol{X_{1}}\right), \\
    & \boldsymbol{X}_{2} = \text{iDFT}\left(\boldsymbol{\mathcal{\hat{F}}}\right), \\
    & \boldsymbol{\mathcal{X}}^{(p)l+1} = \text{Reshape}_\text{i}\left(\boldsymbol{W}_{2} \boldsymbol{X}_{2}\right),
  \end{split}
\end{equation}
where $\boldsymbol{W}_{1} \in \mathbb{R}^{T \times \left(N_{seg}\cdot d_{model}\right)}$ and $\boldsymbol{W}_{2} \in \mathbb{R}^{\left(N_{seg}\cdot d_{model}\right) \times T}$ are learnable parameters for the full series translators and patch translators, a Discrete Fourier Transform (DFT) translator is employed to transform the entire series into the frequency domain, denoted as $\boldsymbol{\mathcal{F}}\in \mathbb{R}^{G\times f}$. Following this, we selectively sample the first half of the spectrum, retaining the portions with the lowest frequency while zeroing out the remainder which is denoted as $\boldsymbol{\mathcal{\hat{F}}}$. Then, we turn to the Inverse Discrete Fourier Transform (iDFT) translator, which transforms the filtered signal back into the temporal domain, denoted as $\boldsymbol{X}_{2}\in\mathbb{R}^{G\times T}$. Lastly, a patch translator is deployed to reshape the denoised full series back into patches $\boldsymbol{\mathcal{X}}^{(p)l+1}\in\mathbb{R}^{G\times N_{seg}\times d_{model}}$ using a linear layer. 

\subsection{Architecture Variants}
%主要总结介绍一下，我的模型的几个不同版本，MHSA-router,MLP-router,MHSA-lin,MHSA-add-attention。后续的主要实验主要对MHSA-router与其他baseline作比较，以及主要对MHSA-router版本做消融实验。
Based on the aforementioned description, SUMformer is more akin to a framework than a specific method. We offer multiple choices in both time and space, as shown in Tab.~\ref{variants}. Note that the main complexity of the attention algorithm is concentrated on the variable dimension since variable number is larger than the input length. The complexity mainly presented in the table is about the variable $G$. We have not listed the computation complexity associated with the input length $T$ in Tab.~\ref{variants}.
\begin{table}[!ht]
\centering
\scriptsize
\caption{Seven Variants of SUMformer}
\label{variants}
\begin{threeparttable}[b]
\begin{tabular}{cccc}
\hline
Variant & TSB & ISSB &  Complexity \\ \hline
SUMformer-AD   & MHSA & NeuralDictionary &  $O(G)$   \\
SUMformer-MD  & MLP & NeuralDictionary &  $ O(G)$  \\
SUMformer-AL  & MHSA & Low-rank Projection &  $O(G)$  \\
SUMformer-AA  &MHSA& Additive Attention &  $O(G)$    \\
SUMformer-AF  & MHSA & Full Attention &   $O(G^2)$  \\
SUMformer-TS  & \multicolumn{2}{c}{NeuralDictionary}  & $O(N_{seg}G)$\tnote{1}\\ 
SUMformer-ViT  & MHSA & NeuralDictionary &   $O\left(\frac{G}{L_{spatial}^2}\right)$  \\\hline
\end{tabular}
\begin{tablenotes}
     \item[1] both space and time.
   \end{tablenotes}
\end{threeparttable}
\end{table} %要不要加一个O(Nseg^2)，其它几种
% \footnotetext[1]{both space and time.}

From Tab.~\ref{variants}, we can find that a model employing full attention is also included. Comparing this with other variants allows us to assess whether the expedited attention mechanism compromises the performance of SUMformer. Of particular note are two distinct variants: SUMformer-TS and SUMformer-ViT. These are instrumental in examining the necessity of temporal attention within SUMformer and evaluating the appropriateness of the super-multivariate perspective for urban mobility data. 

\subsubsection{\textbf{SUMformer-TS}} In the SUMformer-TS variant, we employ the NeuralDictionary mechanism to compute the attention across all variables and patch tokens. This design was intended to investigate the potential of Neural Dictionary, to concurrently capture both temporal and spatial correlations. The fixed value of $g$ and other configuration settings are consistent with those used in SUMformer-AD, resulting in a computational complexity of $O(N_{seg}G)$ for both space and time. 
% It should be noted that we also observed a decline in performance with larger values of $g$ in our evaluations (not listed in our experiments).

%In the first extra experiment, we designed a derivative version of SUMformer which employed one linear attention layer to train all tokens for all variables at every time step denoted as $\boldsymbol{\mathcal{X}}\in\mathbb{R}^{(G\times N_{seg})\times d_{model}}$. 

\subsubsection{\textbf{SUMformer-ViT}} In this variant, we conceptualize urban mobility data akin to a video format by mixing adjacent grids and channels into patch structures. 
Following standard ViTs, we first reshape the input data \( \boldsymbol{\mathcal{X}} \in \mathbb{R}^{T\times C\times H \times W} \) at each time point into a sequence of flattened non-overlapping 2D spatial patches \( \boldsymbol{X} \in \mathbb{R}^{T\times G_{spatial} \times \left(L_{spatial}^2 \cdot C\right)} \), where \( (L_{spatial}, L_{spatial}) \) is the resolution of each spatial patch, and \( G_{spatial} = \frac{HW}{L_{spatial}^2} \) is the resulting number of spatial patches. Subsequently, \(\boldsymbol{X}\) is sliced into non-overlapping temporal sub-series \(\boldsymbol{\mathcal{X}}' \in \mathbb{R}^{G_{spatial}\times N_{seg} \times \left(L_{spatial}^2 \cdot C\cdot L_{seg}\right)} \). Finally, we project the spatio-temporal ``tubes''~\cite{vivit,videoswintransfomer} denoted as \(\boldsymbol{\mathcal{X}}_{i,j}'\in\mathbb{R}^{L_{spatial}^2 \times C\times L_{seg}}\) into 
\(\boldsymbol{\mathcal{X}}_{i,j}^{(p)} \in \mathbb{R}^{d_{model}} \) through a linear embedding layer. The remaining settings align with those of SUMformer-AD.

\section{Experiment}
\label{sec:ex}
We conducted experimental evaluations on real urban mobility datasets spanning three cities: Beijing, Chengdu, and New York. We subsequently delineate various categories of prediction models for comparative analysis, including: the \textbf{Variable-Dependent} Time Series Forecasting Method, the \textbf{Variable-Independent}  Time Series Forecasting Method, the \textbf{Video Prediction} Method, and the \textbf{Frequency-based} Method. Details pertaining to our training procedure are then elaborated upon. Following this, we juxtapose our novel SUMformer architecture with emblematic models from alternative variants, offering an analysis of the relative merits and demerits of each approach. Finally, we present an in-depth analysis of our proposed model.
\subsection{Datasets}
We utilized urban inbound and outbound traffic flow datasets from Beijing, Chengdu, and New York, designated as TaxiBJ, Chengdu, and NYC, for our urban mobility prediction experiments. TaxiBJ and NYC represent taxi flow datasets, while Chengdu pertains to online-hailed vehicle flow. The primary aim is to forecast the inbound and outbound traffic flow for each grid when urban traffic is spatially divided into grids based on latitude and longitude. Detailed attributes for each of the three datasets are presented in Tab.~\ref{dataset_table}. The TaxiBJ and NYC datasets are frequently used in existing literature\footnote{\url{https://github.com/LibCity/Bigscity-LibCity-Datasets}}. In contrast, the Chengdu dataset is proprietary, and we do not have authorization to release it publicly. %用\footnote给出数据集的出处或者下载链接
We primarily use a historical input of 128 frames to predict the traffic flow in the next 128, 64, and 32 frames.
The dataset is split into training, validation, and test sets with a ratio of 7:2:1, respectively.
\begin{table}[!ht]
\centering
\scriptsize
\caption{Statistics of traffic flow datasets}
\begin{tabular}{ccccc}
\hline
Datasets & Spatial size & Variables & Time range & Timesteps \\ \hline
\multirow{4}{*}{TaxiBJ}   & \multirow{4}{*}{$32 \times 32$ }& \multirow{4}{*}{2048} & 2013/7/1--2013/10/29 & \multirow{4}{*}{22484}    \\
~&~&~&2014/3/1--2014/6/27&~\\
~&~&~&2015/3/1--2015/6/30&~\\
~&~&~&2015/11/1--2016/4/10&~\\
Chengdu  & \multirow{1}{*}{$32 \times 32$ } & 2048 & 2023/2/26--2023/7/22 & 7056    \\
NYC      & \multirow{1}{*}{$10 \times 20$ } & 400  & 2015/1/1--2015/3/1  & 2880 \\   \hline
\end{tabular}
\label{dataset_table}

\end{table}

\subsection{Baselines}
We utilized various baseline categories for comparative analysis:
\subsubsection{\textbf{Heuristic Method}}
\begin{itemize}
    \item \textbf{HA}: The \textbf{History Average (HA)} method predicts future values by averaging the input. It represents a widely accepted minimal baseline in existing time series forecasting research.
\end{itemize}

Furthermore, we introduce two enhanced simple baselines to our studies. Urban mobility is renowned for its pronounced daily and weekly periodicities. A high-quality long-term prediction method should ideally surpass both those two methodologies:
\begin{itemize}
    \item \textbf{DH}: The \textbf{Daily History (DH)} method employs historical data from the corresponding time of the previous day for its predictions.
    \item \textbf{WH}: The \textbf{Weekly History (WH)} method employs historical data from the corresponding time of the previous week for its predictions.
\end{itemize}

\subsubsection{\textbf{Video Prediction}}
We compared three video prediction models, both of which are pure convolutional models treating the gird-based data as a video for prediction.
\begin{itemize}
\item \textbf{SimVP}~\cite{simvp} encodes the input frames into a latent space and uses a spatial-temporal translator to learn spatiotemporal variations.
\item \textbf{TAU}~\cite{tau} introduces an attention-based temporal module, extracting inter-frame dynamical attention and intra-frame statistical attention separately.
\item \textbf{Earthformer}~\cite{gao2022earthformer} represents a ViT-based prediction framework, distinct from the CNN-based SimVP and TAU. This method decomposes the video into spatiotemporal 3D cuboids and employs cuboid-level self-attention to discern the spatiotemporal correlations.
\end{itemize}
\subsubsection{\textbf{Variable-Independent Methods}}
We evaluated two time series forecasting techniques grounded in the variable-independent strategy. Such methods approach multivariate time series as a univariate issue, wherein all dimensions utilize shared parameters, overlooking the interrelated correlations between variables.
\begin{itemize}
\item \textbf{Nlinear}\cite{Nlinear} is a one-layer linear model that directly maps the input sequence into the output sequence. To counteract distribution shift, the input is subtracted by the last value of the sequence, passed through a linear layer, and then the subtracted value is added back as a simple form of normalization.
\item \textbf{PatchTST}\cite{patchtst} slices the time series into sub-series level patches, treating them as tokens input to a Transformer.  
%{\color{red} Due to the fact that both PatchTST and our method are based on transformer structures and temporal patches, to ensure a fair comparison, we adjusted the number of encoder layers to 8 while keeping other parameters consistent with the original paper, ensuring that the total parameter count of the models remains similar.} 没必要，各种方法的超参可以写一个附录
\end{itemize}
\subsubsection{\textbf{Frequency-based Methods}}
We also tested several methods that utilize frequency-domain information. Notably, we believe that the spread of human flow in cities may also partially conform to a certain partial differential equations (PDEs)~\cite{schlapfer2021universal}. Precisely because of this reason, we chose the Fourier neural operator (FNO) method, which is highly effective in modeling PDEs, for comparison.
\begin{itemize}
\item \textbf{FNO1D}\cite{FNO} flattens the video into a super-multivariate time series. The historical input data serves as the initial condition for the PDE, and the FNO is employed to solve the PDE, which corresponds to the projected future data.
\item \textbf{FNO3D}\cite{FNO} treats the input video as a whole, and utilize a 3D Fourier transform layer to solve the PDE. 
\end{itemize}

The comparative performance between FNO1D and FNO3D offers an opportunity to evaluate whether the super-multivariate view on urban mobility data is more appropriate than the video-based view.
\begin{itemize}
\item\textbf{Fedformer}\cite{fedformer} integrates a frequency-enhanced block (same as FNO1D), leveraging the seasonal-trend decomposition method to capture the overarching profile of the time series. Concurrently, the Transformer is deployed to discern finer structures. For forecasting, Fedformer employs an implicit variable-dependent strategy.
\end{itemize}
\subsubsection{\textbf{Variable-Dependent Methods}}
\begin{itemize}
    \item \textbf{TCN}\cite{TCN} utilizes a 1-D causal convolution to ensure the model relies only on past information during predictions, while simultaneously implicitly modeling correlations across variable dimensions.
    \item \textbf{Crossformer}\cite{crossformer} devised a Two-Stage-Attention (TSA) layer to capture the cross-time and cross-variable dependency. A hierarchical encoder-decoder architecture is employed for multi-scale information utilization.
\end{itemize}

For the time series forecasting benchmarks, the variables across all channels and grids are evaluated as distinct time series.

\subsection{Experiment Settings}
Our experiments were conducted on a server equipped with a 20-core Intel Silver 4316 CPU and an NVIDIA GeForce RTX 4090 GPU. The SUMformer model was implemented using PyTorch, and the implementation code has been released on GitHub\footnote{\url{https://github.com/Chengyui/SUMformer}}. The dataset was divided into training, validation, and testing sets with a ratio of 7:1:2. We trained eight baselines and SUMformer for 80 epochs with a batch size of 16. The Adam optimizer was used, and the learning rate was scheduled using the CosineLRScheduler from the \textit{timm} library. The warm-up phase consisted of 5 epochs, with the learning rate set to 1e-5, and in the training phase, the learning rate was set to 5e-4.

The primary hyperparameter settings for SUMformer are as follows: we utilize SUMformer-AD in Tab.~\ref{bigtable}, which employs MHSA in the temporal domain and a neural dictionary for the spatial domain. The patch merge ratio, \(r_{\text{win}}\), is set to 2. The initial length of the sub-series, \(L_{\text{seg}}\), is 16, while the fixed dictionary size for spatial linear attention is 256. The embedding size, \(d_{\text{model}}\), is 128. Additionally, the number of TVF blocks is 4, indicating that the patch will undergo merging four times. We employed a grid search approach based on the default settings of other baselines reported in the literature, ensuring that these approaches achieved optimal performance. We evaluated each model by root-mean square error (RMSE) and mean absolute error (MAE).

\subsection{Main results}

% 主要结论：super multivariate view > video view (FNO1D > FNO3D, our method > others, Xudong's paper); explict dependencies modeling is important, but the video way is not good (our and crossformer > others)
 \begin{table*}[!htb]
     \centering
     \scriptsize
     \renewcommand{\arraystretch}{1.2}
     \caption{MAE/RMSE comparision on three datasets}
     \begin{tabular}{lll|ll|ll|ll}
     \hline
         Dataset & ~ & Scenario & \multicolumn{2}{c}{128-128}  & \multicolumn{2}{c}{128-64} & \multicolumn{2}{c}{128-32} \\ \hline
         ~ &  Method categories & Method & MAE & RMSE & MAE & RMSE & MAE & RMSE \\ \hline
         \multirow{12}{*}{TaxiBJ}    & \multirow{3}{*}{Heuristic Method} &HA &43.023 &70.321 &43.133 &70.488 &43.168 &70.541\\
         ~ & ~ & WH & 36.402 & 71.085 & 36.402 & 71.085 & 36.402 & 71.085 \\ 
         ~ & ~ & DH & 23.425 & 47.472 & 23.425 & 47.472 & 23.425 & 47.472 \\ \cline{2-9}
         ~ & \multirow{3}{*}{Frequency-based} & FNO3D~\cite{FNO} & 26.019 & 47.151 & 22.038 & 39.911 & 19.833 & 35.317 \\ 
         ~ & ~ & FNO1D~\cite{FNO} & 25.422 & 47.595 & 21.057 & 38.481 & 19.101 & 34.746 \\ 
        ~ & ~ & Fedformer~\cite{fedformer} & 24.480 & 46.255 & 21.244 & 38.820 & 19.288 & 35.546 \\ \cline{2-9}
         ~ & \multirow{3}{*}{Video Prediction}& SimVP~\cite{simvp} &  22.154 & 40.764 & 19.246 &  36.311 & 16.375 & \underline{30.676} \\ 
         ~ & ~ & TAU~\cite{tau} & 23.082 & 42.177 & 19.105 & 36.644 & 17.422 & 32.921 \\ 
         ~ & ~ & Earthformer~\cite{gao2022earthformer} & 25.816 & 48.435 & 20.637 &38.386 & 17.106 &32.655 \\ \cline{2-9}
         ~ & \multirow{2}{*}{Variable-Independent} & 
         Nlinear~\cite{Nlinear} & 25.778 & 47.801 & 21.393 & 40.411 & 19.306 & 36.899 \\ 
         ~ & ~ & PatchTST~\cite{patchtst} & \underline{20.674} & \underline{40.597} & \underline{18.291} & 34.933 & \underline{16.166}& 31.330 \\ \cline{2-9}
         ~ & \multirow{3}{*}{Variable-Dependent} & 
         TCN~\cite{TCN} & 25.004 & 46.787 & 21.503 & 38.693 & 18.912 & 35.076 \\ 
         ~ & ~ & Crossformer~\cite{crossformer} & 21.342 &40.732 & 18.784 & \underline{34.626} & 19.765 &33.553 \\ 
\rowcolor{gray!20}         ~ & ~ & SUMformer &\textbf{19.347} & \textbf{38.102} & \textbf{16.681} & \textbf{32.619} & \textbf{15.268} & \textbf{29.815 }\\ \hline
                 \multirow{12}{*}{Chengdu}    & \multirow{3}{*}{Heuristic Method} &HA &11.449&26.737 &11.425 &26.684 &11.384 &26.602\\
        ~ & ~ & WH & 4.543 & 10.512 & 4.543 & 10.512 & 4.543 & 10.512 \\ 
         ~ & ~ & DH & 4.549 & 10.762 & 4.549 & 10.762 & 4.549 & 10.762 \\ \cline{2-9}
         ~ & \multirow{3}{*}{Frequency-based} & FNO3D~\cite{FNO} & 4.555 & 10.426 & 4.553 & 10.020 & 4.388 & 9.607 \\ 
         ~ & ~ & FNO1D~\cite{FNO} & 4.011 & 9.551 & 3.841 & 8.915 & 3.789 & 8.839 \\ 
         ~ & ~ & Fedformer~\cite{fedformer} & 4.370 & 9.885 & 4.170 & 9.313 & 4.406 & 9.287 \\ \cline{2-9}
         ~ & \multirow{3}{*}{Video Prediction}& SimVP~\cite{simvp} & 4.559 & 10.160 & 4.437 & 9.600 & 4.108 & 8.904 \\ 
         ~ & ~ & TAU~\cite{tau} & 4.599 & 9.632 & 4.286 & 9.373 & 4.118 & 8.856 \\ 
         ~ & ~ & Earthformer~\cite{gao2022earthformer} & 4.309 & 10.302 & 4.187 & 10.150 & 4.142 & 9.744 \\ 
                 \cline{2-9}
         ~ & \multirow{2}{*}{Variable-Independent} & 
          Nlinear~\cite{Nlinear} & 4.383 & 10.800 & 4.184 & 10.197 & 4.065 & 9.796 \\ 
         ~ & ~ & PatchTST~\cite{patchtst} & 4.015 & 9.424 & 3.824 & 8.988 & 3.710 & 8.629 \\ \cline{2-9}
         ~ & \multirow{3}{*}{Variable-Dependent} & 
         TCN~\cite{TCN} & 4.369 & 10.691 & 4.213 & 9.900 & 4.082 & 9.669 \\ 
         ~&~&Crossformer~\cite{crossformer} &\underline{3.997} & \underline{9.203} & \underline{3.816} & \underline{8.578} & \underline{3.683} & \underline{8.082} \\ 
\rowcolor{gray!20}         ~&~ &SUMformer & \textbf{3.789} &\textbf{8.830} & \textbf{3.728} & \textbf{8.377} & \textbf{3.678} & \textbf{7.997} \\ \hline
                 \multirow{12}{*}{NYC}    & \multirow{3}{*}{Heuristic Method} &HA &20.395&58.328&20.370&58.382&20.520&58.743\\
         ~ & ~ & WH & 7.111 & 22.298 & 7.111  & 22.298 &  7.111 & 22.298 \\ 
         ~ & ~ & DH & 10.101 & 35.694 &10.101~ & 35.694 & 10.101 & 35.694 \\ \cline{2-9}
         ~ & \multirow{3}{*}{Frequency-based} & FNO3D~\cite{FNO} & 9.247 & 23.926 & 8.040 & 20.552 & 7.716 & 19.343 \\ 
         ~ & ~ & FNO1D~\cite{FNO} &6.176 & \underline{17.517} & 6.027 & 16.886 & 5.816 & 15.964 \\ 
        ~ & ~ & Fedformer~\cite{fedformer} & 7.644 & 22.379 & 7.420 & 20.987 & 6.570 & 18.681 \\ \cline{2-9}
         ~ & \multirow{3}{*}{Video Prediction}& SimVP~\cite{simvp} & 6.978 & 19.014 & 6.141 & 16.666 & 5.859 & 16.096 \\ 
         ~ & ~ & TAU~\cite{tau} & 7.298 & 18.993 & 6.430 & 17.437 & 5.899 & 16.328 \\ 
        ~ & ~ & Earthformer~\cite{gao2022earthformer} & 6.465 & 17.848 & 5.846 & 17.455 & 5.882 & 17.014 \\ 
         \cline{2-9}
         ~ & \multirow{2}{*}{Variable-Independent} & 
        Nlinear~\cite{Nlinear} & 12.072 & 38.261 & 10.593 & 34.118 & 9.653 & 31.511 \\ 
        ~ & ~ & PatchTST~\cite{patchtst} & 7.488 & 22.656 & 6.987 & 20.706 & 6.581 & 18.708 \\ \cline{2-9}
        ~ & \multirow{3}{*}{Variable-Dependent} & 
        TCN~\cite{TCN} & 8.766 & 22.622 & 7.599 & 19.662 & 8.576 & 19.484 \\ 
        ~&~& Crossformer~\cite{crossformer} &\underline{6.138} & 17.592 & \underline{5.696} & \underline{16.346} &\underline{5.534} & \underline{15.537} \\ 
\rowcolor{gray!20}         ~&~ & SUMformer & \textbf{5.595} &\textbf{16.734} & \textbf{5.389} & \textbf{15.607 }& \textbf{5.121 }&\textbf{14.879} \\ \hline
     \end{tabular}
     \label{bigtable}
 \end{table*}

The comprehensive results are presented in Tab.~\ref{bigtable} in which lower MAE and RMSE values indicate more accurate predictions, with the optimal results highlighted in \textbf{bold} and the second-best results \underline{underlined}. Our proposed SUMformer model consistently outperforms across all scenarios and metrics. Compared to video prediction methods, SUMformer exhibits a significant performance advantage by a large margin. Besides demonstrating the superiority of SUMformer, the results in Tab.~\ref{bigtable} also provide a comprehensive comparison of different types of methods, including History, Frequency-based, variable independent, variable dependent, as well as a comparison between explicit and implicit modeling methods related to cross-variable correlation. We can derive three key observations from these results:

\textbf{\textit{1). Super-multivariate Modeling Outperforms Video Modeling in Long-term Urban Mobility Forecasting:}} For Chengdu and TaxiBJ datasets, which presumably have larger training data sizes, the strong time series forecasting baselines - PatchTST, Crossformer, and our SUMformer - consistently outshine the video-based prediction models such as SimVP, TAU, and Earthformer. Notably, even for the NYC dataset, both Crossformer and SUMformer continue to surpass the video-based forecasting models. This assertion is further bolstered when comparing the performance of FNO1d and FNO3d, where the former consistently delivers better results,  

\textbf{\textit{2). Explicit Variable Correlation Modeling is Crucial:}} We included the Fedformer baseline, which translates the variable dimension into a latent space for implicitly modeling cross-variable relationships. Despite being equipped with a frequency-enhancement block, this method doesn't yield impressive results. PatchTST, which doesn't even focus on cross-variable correlation, still surpasses Fedformer. In the majority of scenarios and across various metrics, both PatchTST and Fedformer lag behind Crossformer and our proposed SUMformer. This underscores the importance of explicitly modeling variable correlations when dealing with urban mobility forecasting. Our SUMformer, armed with an attention mechanism that boasts linear computational complexity, excels in both efficiency and effectiveness when explicit modeling cross-variable correlations.

\textbf{\textit{3). Emphasizing the Periodicity of Urban Mobility is Crucial for Long-term Forecasting:}} We presented two robust historical benchmarks, WH and DH, tailored for extended-horizon forecasting. Intriguingly, though these methods rely primarily on data from the previous day and week for their predictions, they manage to outshine several advanced deep learning baselines, most notably in the longest forecasting window (128-128). This phenomenon can be attributed to the innate statistical properties of urban mobility. Research indicates that as the time lag extends, the autocorrelation coefficient of urban mobility wanes~\cite{wang2023anti}. It underscores the importance of prediction frameworks that harness the intrinsic periodic, seasonal, and recurrent patterns in the data for long-term accuracy. Our SUMformer model, with its low-frequency filtering mechanism, adeptly captures and emphasizes these periodic patterns.

\begin{figure*}[t]
\centering    
{\includegraphics[width=1\textwidth]{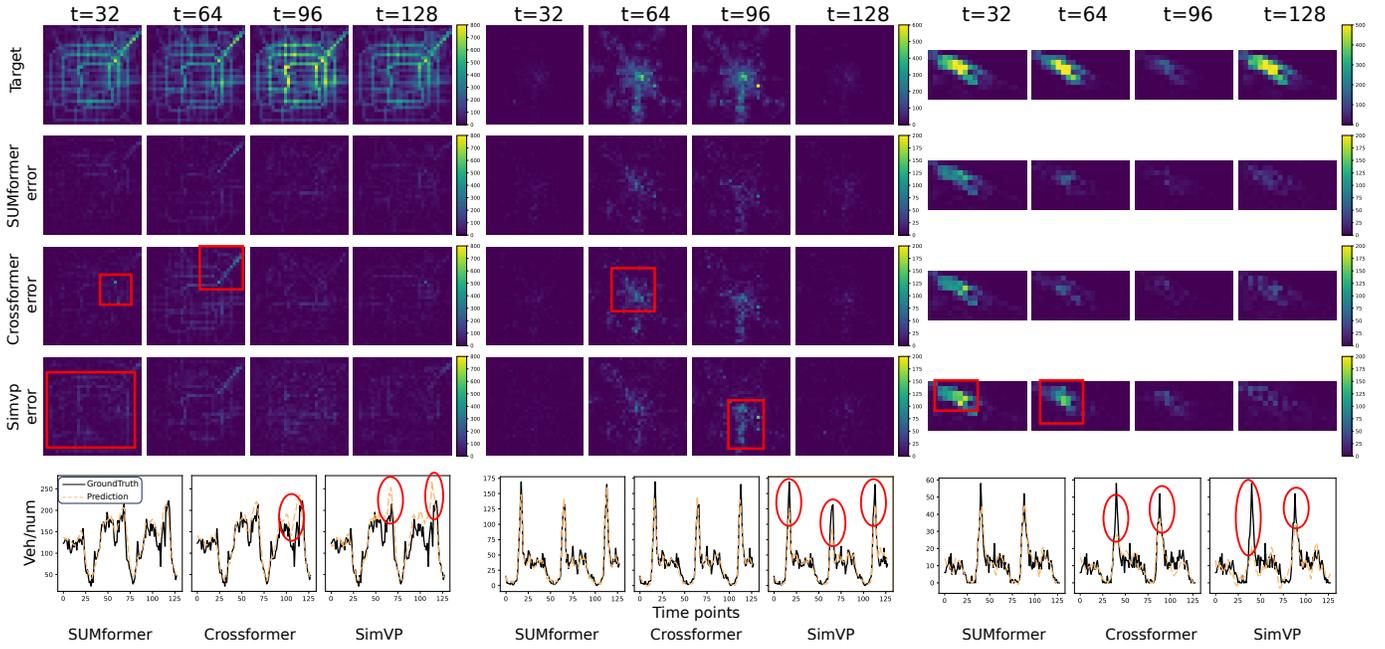}}
\caption{Spatial and temporal visualization for SUMformer, Crossformer, and SimVP on TaxiBJ, Chengdu, NYC respectively while the two channels are averaged for convenience. Significant errors are highlighted with red boxes.}
\label{space_vis}
\end{figure*}

We also visualize predicted errors of SUMformer, Crossformer and SimVP on three datasets in Fig.~\ref{space_vis}. At various prediction points, labeled as $t=32$, $t=64$, $t=96$, and $t=128$, the target displays intricate patterns. In addition, we provide the outputs of the predicted time series from three methods for a specific grid within each dataset. We have observed that human mobility data exhibit a strong periodic pattern, and all three methods can offer predictions that are fundamentally accurate on a macro trend level. SUMformer's errors show minimal deviations from the target, suggesting a high level of accuracy. In contrast, both Crossformer and SimVP appear to have more pronounced error patterns, indicating larger discrepancies from the target. This visualization underscores the superior accuracy and precision of SUMformer in comparison to the other two models. SimVP performs the poorest, particularly in the NYC example, suggesting that CNN-based methods may not be suitable for long-term prediction tasks. SUMformer's advantage over Crossformer can be attributed to its use of low-frequency filters and a more sophisticated Swin Transformer-style architecture. It tracks larger values more accurately than Crossformer in those three examples. In Fig.~\ref{space_vis}, we selected one variable from each of the three datasets for temporal visualization. We observed that SUMformer can effectively capture the temporal patterns of traffic flow, particularly in predicting the peak traffic flow, which is a crucial metric for traffic agencies.
\subsection{Performance Analysis for Seven Variants}

\begin{table}[!ht]
    \centering
    \scriptsize
    \renewcommand{\arraystretch}{1.2}
    \caption{MAE/RMSE comparision on Seven variants}
    \begin{tabular}{ll|ll|ll|ll}
    \hline
        Dataset & Scenario & \multicolumn{2}{c}{128-128}  & \multicolumn{2}{c}{128-64} & \multicolumn{2}{c}{128-32} \\ \hline
        ~  & Variant & MAE & RMSE & MAE & RMSE & MAE & RMSE \\ \hline
        \multirow{7}{*}{TaxiBJ} &AD & \underline{19.347} & \textbf{38.102} &\textbf{16.681} & 32.619 & 15.268 & 29.815\\
        ~ & MD & 20.091 & 39.364 & 17.403 & 33.542 & 15.733 & 30.503 \\ 
        ~ & AL & \textbf{19.138} & \underline{38.361} & 17.169 & \textbf{32.498} & \textbf{14.915} & \underline{28.883} \\ 
        ~ & AA & 20.442 &39.384 & 17.628 &34.009 & 15.774 & 30.133 \\ 
        ~ & AF & 19.890 & 38.801 & \underline{16.765} & \underline{32.664} & \underline{15.109} & \textbf{28.878} \\ 
        ~ & TS & 21.416 & 41.222 & 17.797 & 34.033 & 16.242 & 31.432 \\ 
        ~& ViT&20.526 &40.619 &17.586 &33.126 &15.504 &29.085 \\
        \cline{2-8}
        \hline
        \multirow{7}{*}{Chengdu} &AD &\textbf{3.789} &8.830 &\underline{3.728} &\textbf{8.377} &\textbf{3.678} &\textbf{7.997}\\
        ~ & MD & 3.862 & \underline{8.826} & 3.840 & 8.773 & 3.778 & 8.281 \\ 
        ~ & AL & 4.040 & 9.471 & 3.887 & 9.036 & 3.914 & 8.832 \\ 
        ~ & AA & 3.966 & 9.120 & 3.857 & 9.113 & 3.930 & 8.562  \\ 
        ~ & AF & 3.958 & 8.866 & 3.806 & 8.848 & \underline{3.710} & \underline{8.204} \\ 
        ~ & TS & \underline{3.851} & \textbf{8.765} & \textbf{3.724} & \underline{8.380} & 3.790 & 8.287 \\ 
        ~& ViT&4.132 &9.326 &3.988 & 9.036&3.845 &8.576 \\
        \cline{2-8}
        \hline
        \multirow{7}{*}{NYC} &AD & \underline{5.595} & \underline{16.734} & 5.389 & \textbf{15.607} &\textbf{5.121} & \textbf{14.879}\\
        ~ & MD & 5.768 & 17.465 & 5.405 & 16.158 & \underline{5.178} & 15.127 \\ 
        ~ & AL & \textbf{5.561} & \textbf{16.631} & \underline{5.338} & 15.805 & 5.256 & 15.414 \\ 
        ~ & AA & 5.817 & 17.834 & 5.559 & 16.172 & 5.640 &15.425 \\ 
        ~ & AF & 5.795 & 17.276 & \textbf{5.318} & \underline{15.668} & 5.231 & 15.161 \\ 
        ~ & TS & 6.497 & 19.179 & 5.684 & 16.560 & 5.350 & \underline{15.053}  \\
        ~& ViT &6.235 &17.909 &5.715 & 16.646 &5.439  &15.532 \\
        \hline
    \end{tabular}
    \label{six_variants}
\end{table}

\begin{figure}[htp]
\centering    
{\includegraphics[width=0.45\textwidth]{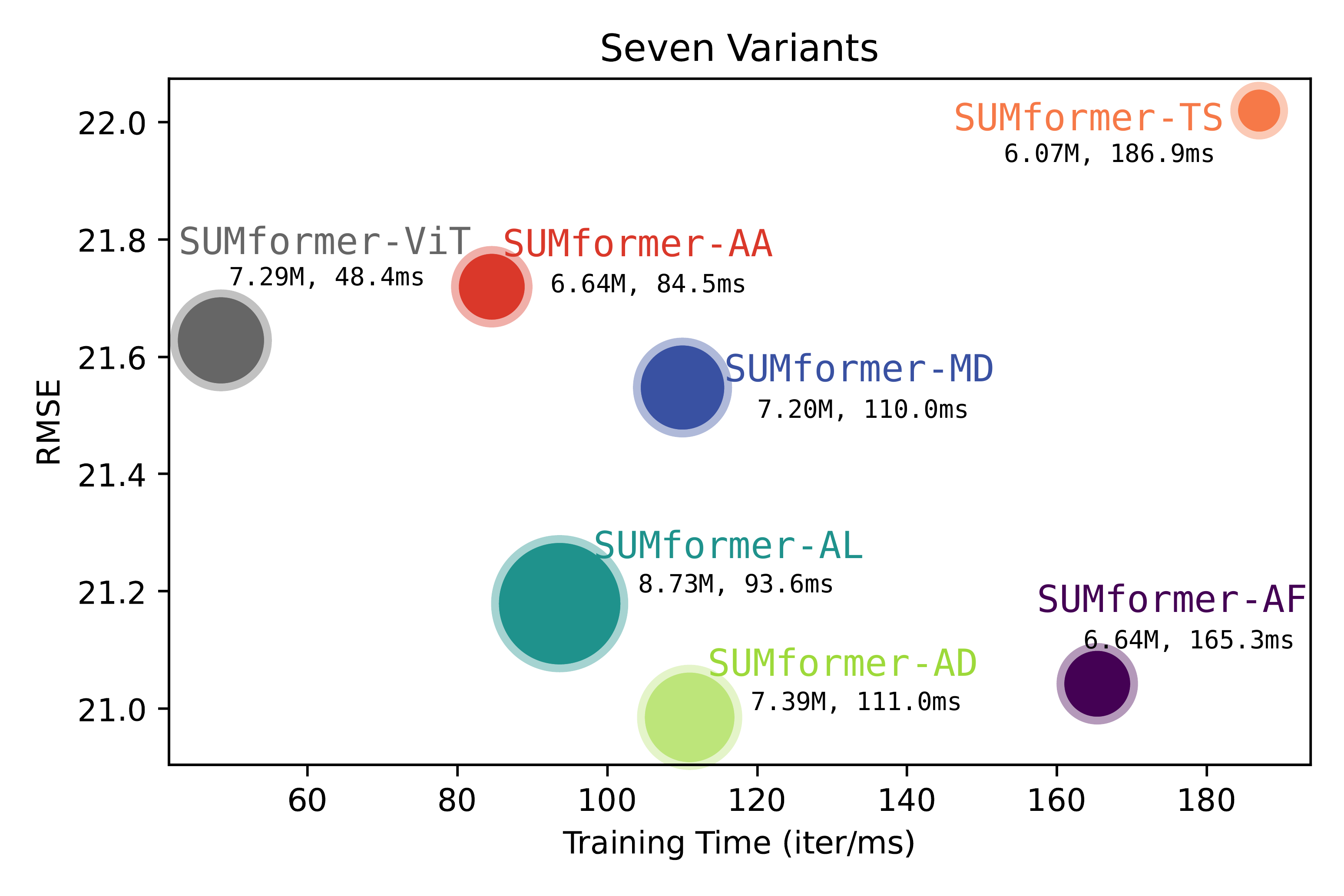}}
\caption{Average model size, training speed, and performance for seven variants. Their sizes are proportional to their areas.}
\label{Param_size}
\end{figure}

%We analyze different variants of SUMformer in this section. Firstly, we examine the performance of six variants on three datasets. Subsequently, we investigate the impact of different spatial patch sizes, $L_{spatial}$, on the performance of SUMformer-ViT when modeling traffic flow as a video.
Tab.~\ref{six_variants} shows the results of seven variants on the three datasets, where SUMformer-ViT uses a $2\times 2$ patch. We provide a comprehensive comparison of training speed, model size, and performance on TaxiBJ and Chengdu in Fig.~\ref{Param_size}. Our key observations (Obs) are as follows:

\textbf{\textit{Obs 1):}} SUMformer-AD and SUMformer-AL consistently achieve the best performance across most scenarios. This underscores that linear attention techniques—whether leveraging a neural dictionary or linear layers to produce low-rank key-value pairs—maintain high accuracy while reducing computational demands. In many instances, their results even outshine those of SUMformer-AF, which incorporates full attention. A possible explanation is that \textbf{the correlation between the urban mobility data is inherently low-rank}, and full attention models might face challenges in grasping this structure. The suboptimal results of SUMformer-AA, which uses additive attention, suggest that utilizing a global vector via additive attention might not be the optimal approach for capturing the low-rank correlations among variables. \textbf{\textit{Obs 2):}} SUMformer-MD, which leverages MLP within the Temporal Sub-Block to extract temporal correlations, underperforms compared to SUMformer-AD. This implies that MHSA is more effective than MLP at extracting temporal correlations. \textbf{\textit{Obs 3):}} SUMformer-TS, which omits the Temporal Sub-Block and relies solely on the Neural Dictionary to extract both temporal and cross-variable correlations, is the slowest method and yields the poorest prediction results. While SUMformer-ViT is the fastest method, thanks to its avoidance of computing attention scores for all variables, its prediction accuracy lags behind most other variants, with the exception of SUMformer-TS and SUMformer-AA. This suggests that a two-stage approach to modeling temporal and cross-variable correlations step-by-step is well-suited for urban mobility data.

\subsection{Ablation Study}
\begin{figure}[htp]
\centering    
{\includegraphics[width=0.45\textwidth]{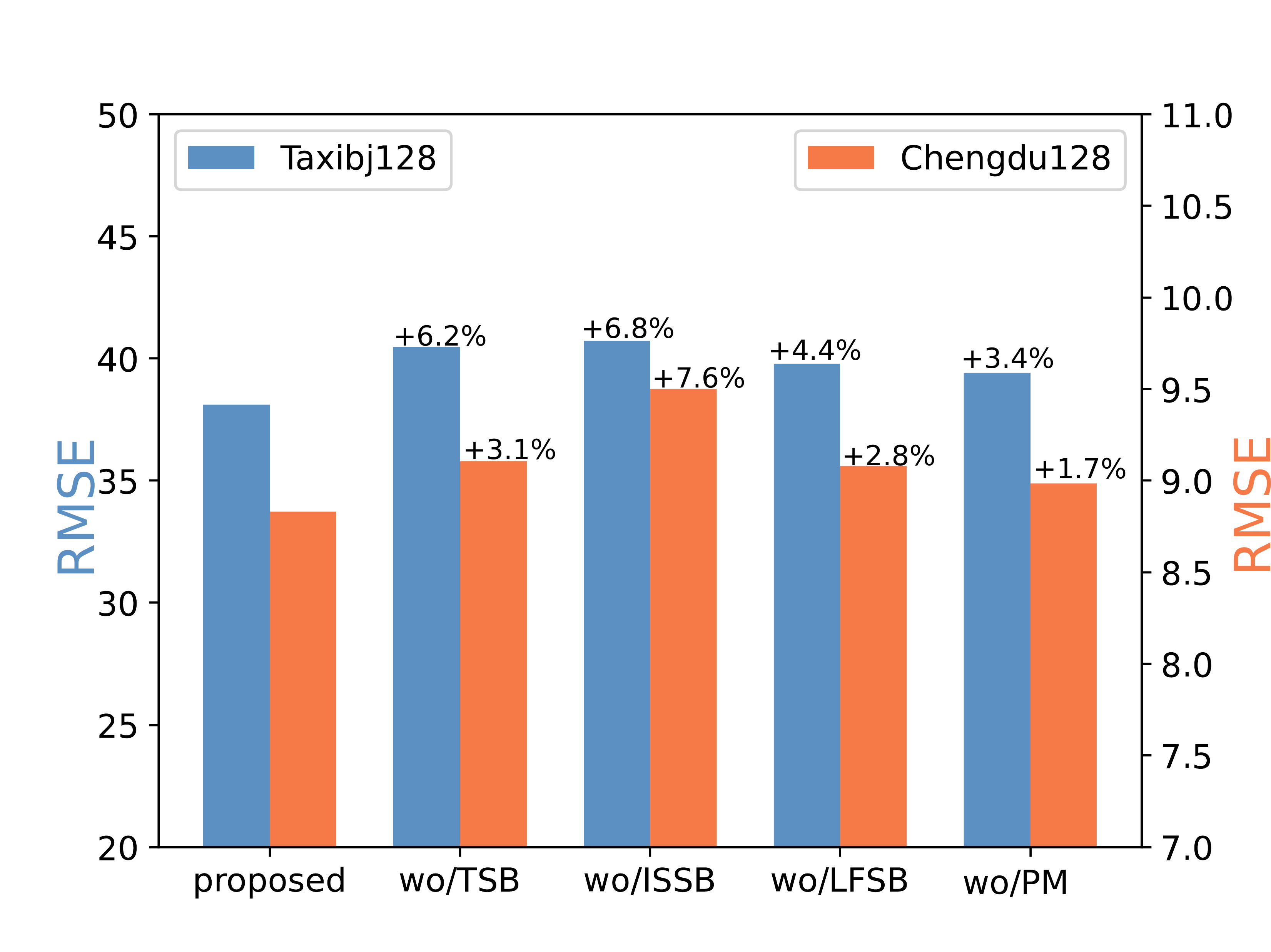}}
\caption{Main component ablation for SUMformer}
\label{bar}
\end{figure}

We conducted ablation testing to assess the effectiveness of the SUMformer architecture. We explored four ablation strategies on \textbf{TSB}, \textbf{ISSB}, \textbf{LFSB} as well as \textbf{Patch Merge mechanism} and assessed their impact on 128-step-ahead predictions using the TaxiBJ and Chengdu datasets.

% We compared three models. First, we removed LFSB to create \textbf{TSB+PM+ISSB}. After that, by removing ISSB, we obtained \textbf{TSB+PM}. Finally, by removing PM, we were left with a pure temporal Transformer, \textbf{TSB}. 
Fig.~\ref{bar} presents the results of the ablation study. All experiments were conducted using the SUMformer-AD as the baseline for evaluation. 
In summary, each component of the SUMformer\textemdash TSB, ISSB, LFSB and the Patch Merge mechanism, holds significant importance in ensuring accurate predictions. The ablation study highlights the critical importance of each component. Omitting any of them consistently results in increased prediction errors, with the ISSB standing out as the most crucial. We observed a 6.8\% increase in RMSE for TaxiBJ and a 7.6\% increase for Chengdu. This underscores the significance of modeling inter-series correlations within urban mobility data. The ISSB allows the model to explicitly capture correlations among variables. This ensures that predictions are not just based on individual variables but also encompass a comprehensive understanding of global patterns. Furthermore, our observations indicate that removing the TSB results in a less significant performance reduction compared to eliminating the ISSB. This indicates that the model is capable of conducting implicit temporal modeling through the synergistic effects of ISSB, LFSB, and particularly the patch merge mechanism. It's plausible that the main contribution to this capability arises from the patch merge mechanism, which merges temporally neighboring patches using linear layers, thereby implicitly modeling temporal correlations.

%The rich cross-variable correlations play a crucial role, and neglecting these connections would result in a significant degradation of forecasting accuracy. ISSB enables the model to capture the correlations among variables explicitly, ensuring predictions are not solely reliant on individual variables and encompass a holistic understanding of the global features.

%This could be attributed to the nature of long-term predictions, which require the model to comprehend both current and historical states. The Patch Merge mechanism enables the model to capture the temporal continuity of the data, ensuring predictions aren't solely based on recent events but encompass a comprehensive understanding of historical patterns.

\subsection{Effect of hyper-parameter}

\begin{figure}[t]
\centering    
\subfigure[Effect of dictionary dimension $g$ for SUMformer-AD]{
\label{ablation_g}
\includegraphics[width=0.46\linewidth]{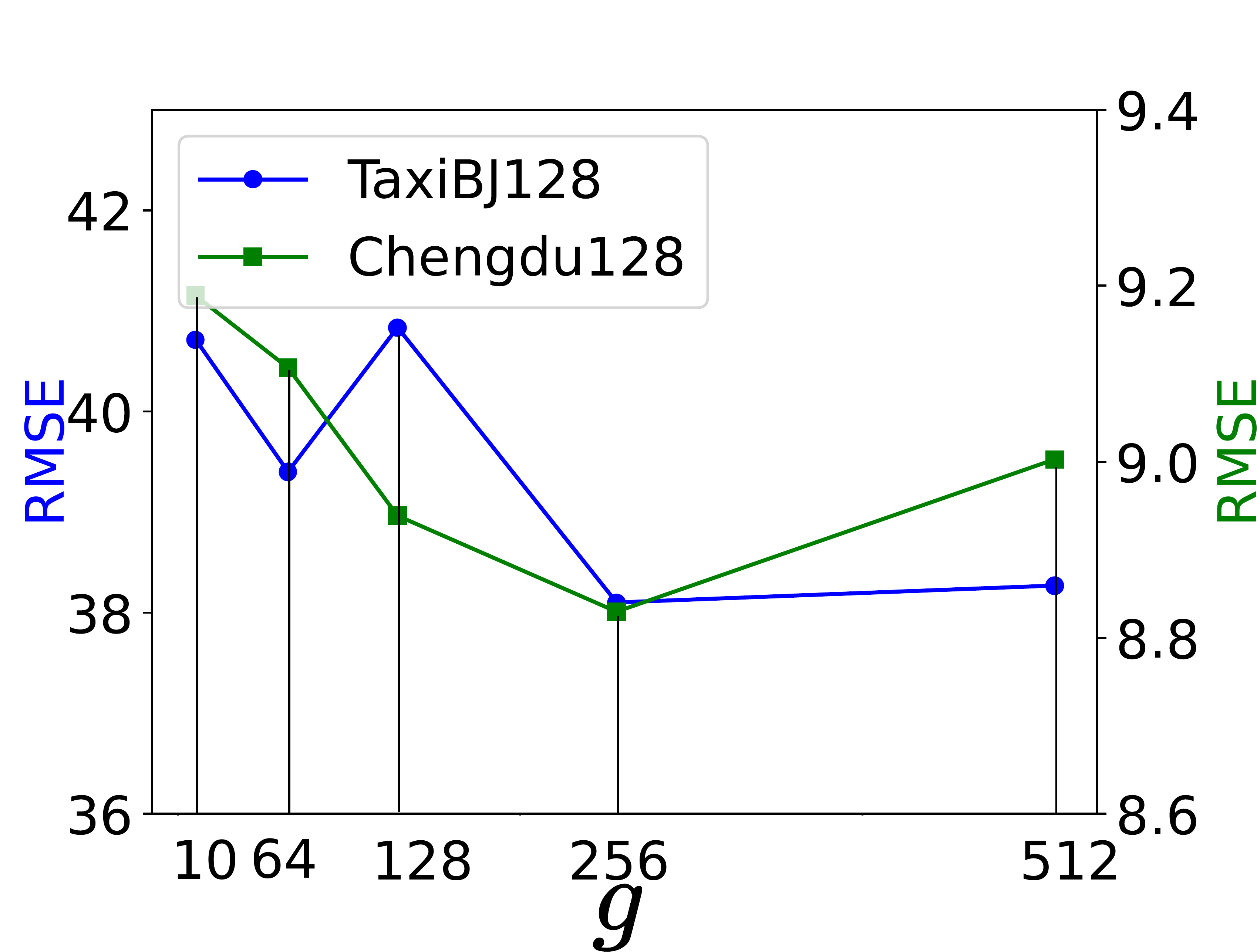}}
\subfigure[Effect of input length $T$]{
\label{ablation_T}
\includegraphics[width=0.46\linewidth]{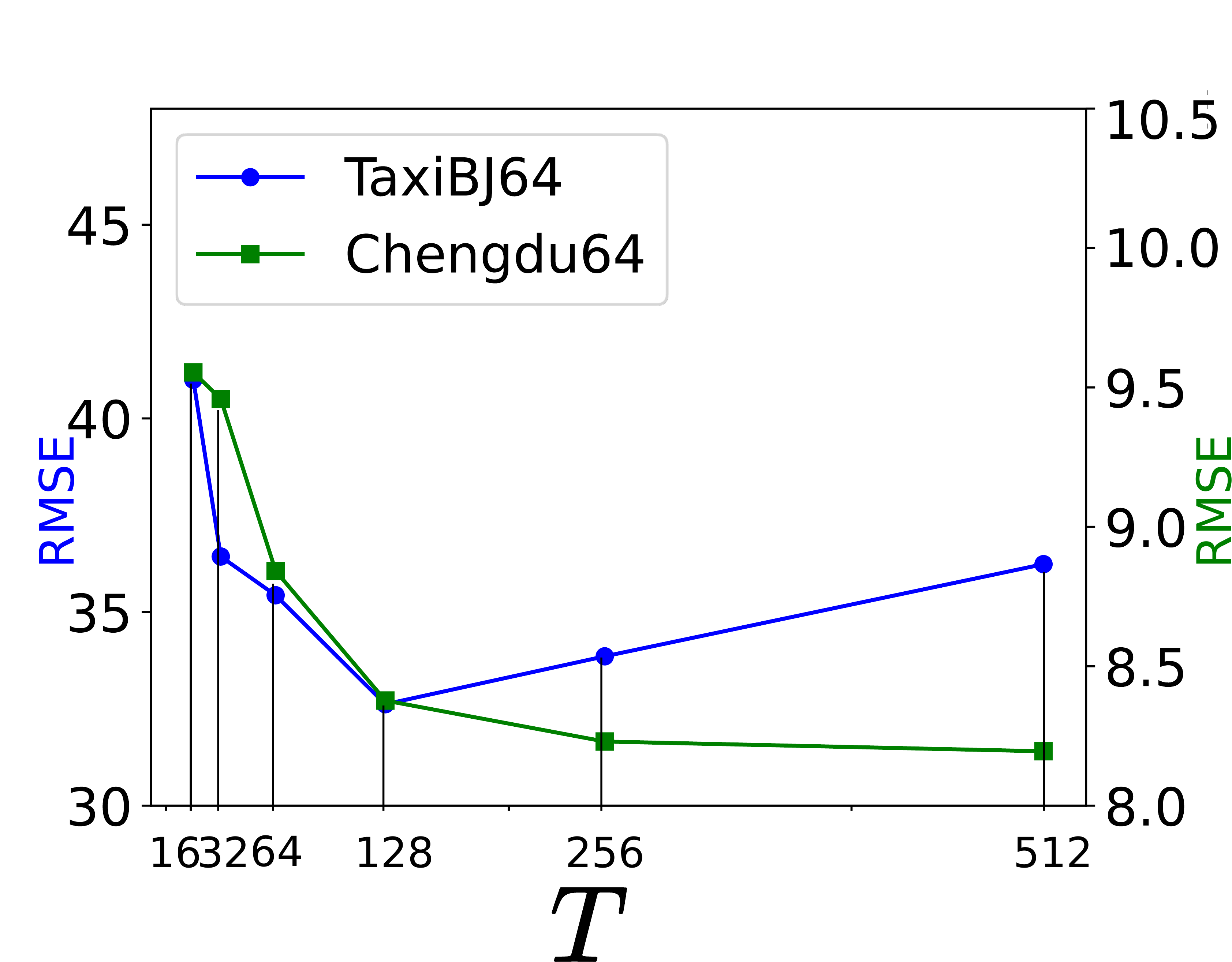}}
\subfigure[Effect of the length of the patch merged window $r_{win}$]{
\label{ablation_rwin}
\includegraphics[width=0.46\linewidth]{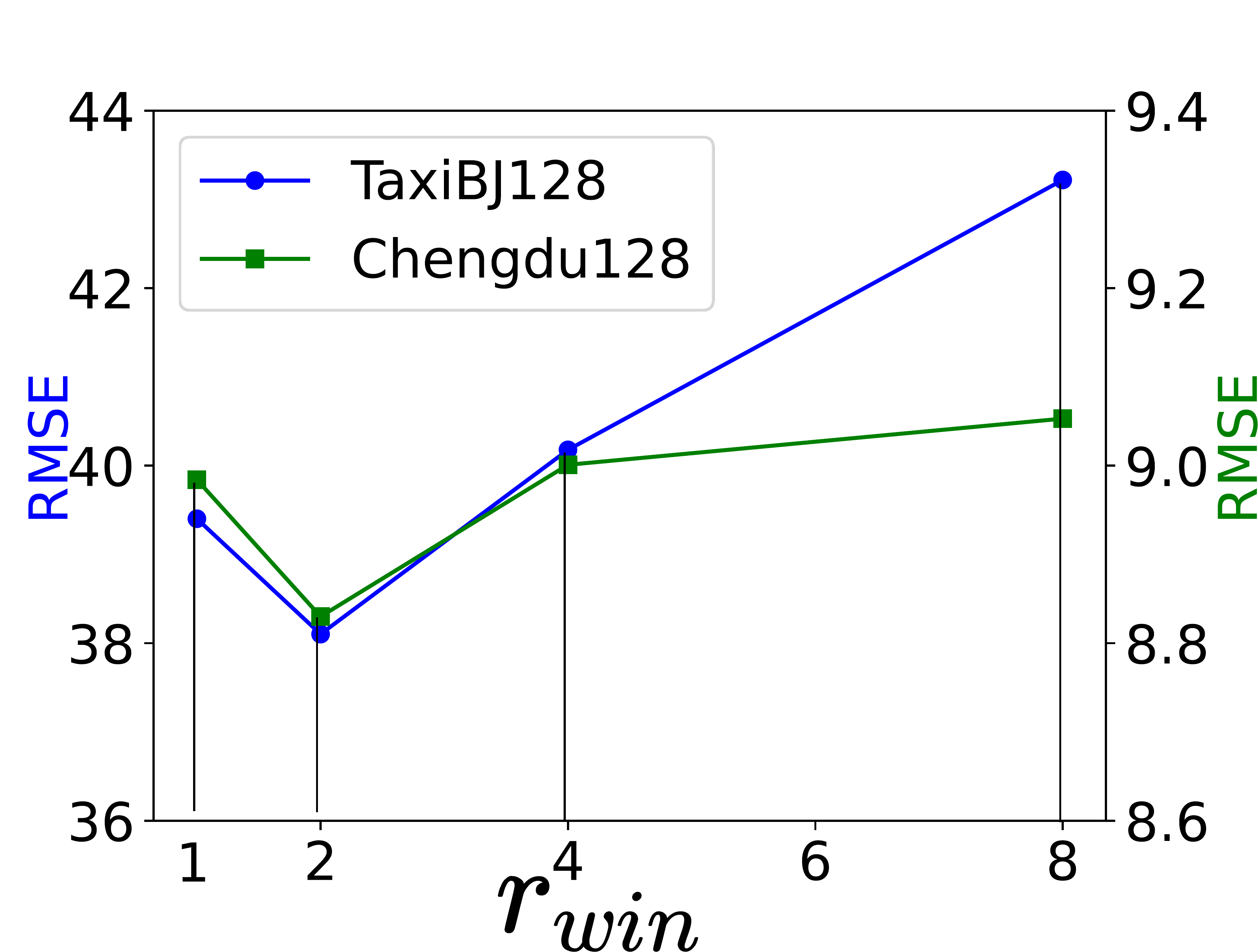}}
\subfigure[Effect of spatial patch size $L_{spatial}$ for SUMformer-ViT]{
\label{ablation_lspatial}
\includegraphics[width=0.46\linewidth]{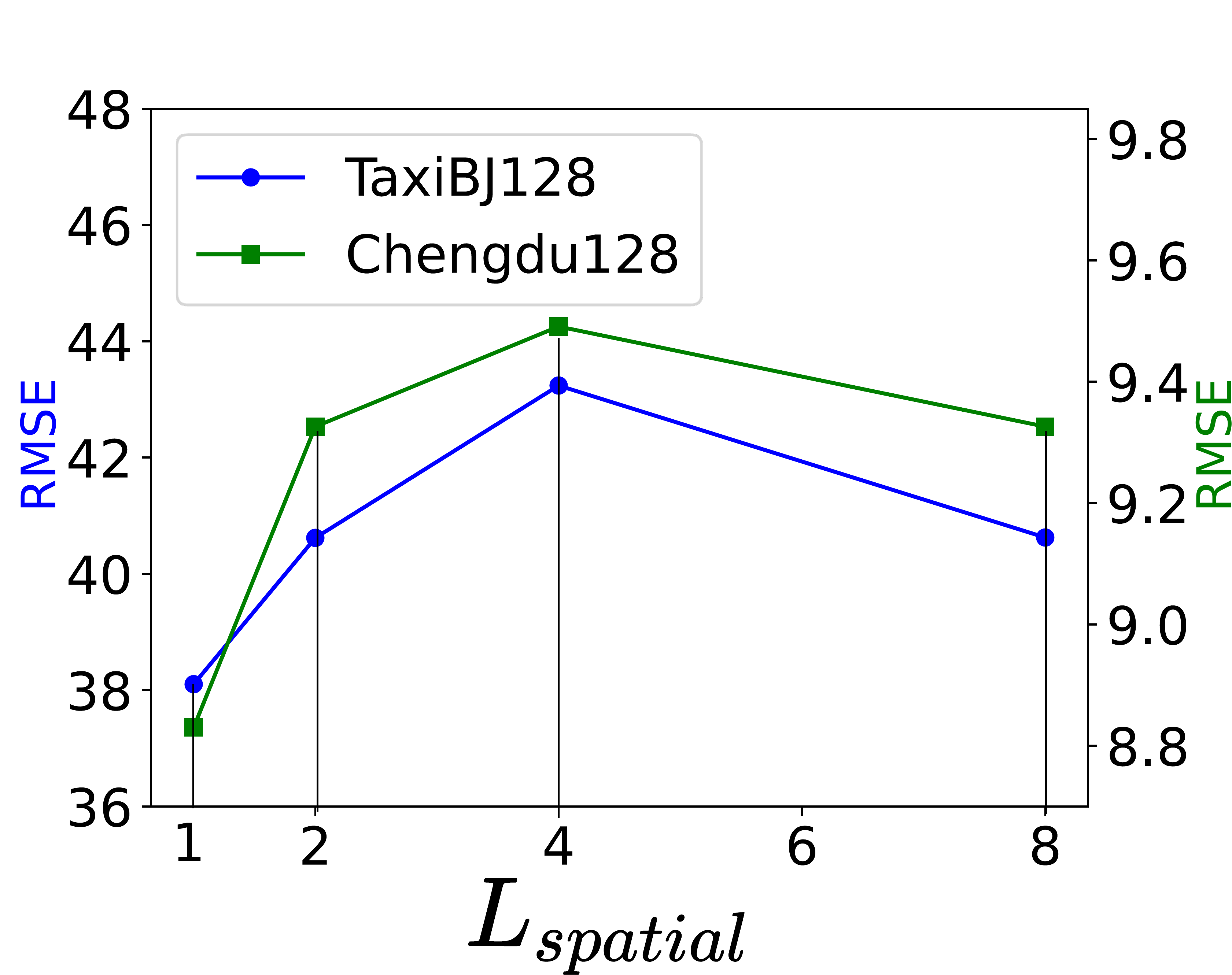}}
\caption{Evaluation of hyper-parameter impact on prediction accuracy.}
\label{hyper_param}
\end{figure}

\subsubsection{\textbf{Selection of dictionary dimension $g$ for SUMformer-AD}} Fig.~\ref{ablation_g} shows the forecasting accuracy, measured in terms of the Root Mean Square Error (RMSE), for a 128 step-ahead prediction on both the TaxiBJ and Chengdu datasets using the SUMformer-AD model with varying dictionary dimensions. Two datasets, TaxiBJ and Chengdu, are plotted using blue and green lines, respectively.  
For both datasets, a dictionary dimension of \(g = 128\) appears to yield the best forecasting accuracy. Lower dimensions, particularly \(g = 10\), lead to inferior performance, indicating a requisite threshold for the dictionary size to capture cross-variable correlation effectively. Notably, while the performance for TaxiBJ diminishes considerably for larger \(g\) values such as 256 and 512, the Chengdu dataset maintains a relatively consistent performance, highlighting inherent differences in the data characteristics of the two datasets. Note that both datasets contain over 2,000 variables. The fact that the model with 
$g=256$ achieves the best performance suggests that both datasets exhibit low-rank cross-variable correlations.

\subsubsection{\textbf{Performances under Longer Input Sequences}} A proficient time series forecasting model should accurately capture dependencies over extended review windows, thereby enhancing its results. A previous study~\cite{Nlinear} showed that Transformer-based models can exhibit significant fluctuations in performance, resulting in either a decline in overall efficiency or decreased stability with longer review windows. We conducted a similar analysis on both the TaxiBJ and Chengdu datasets using the SUMformer-AD model. Various input lengths, specifically $\{16, 32, 64, 128, 256, 512\}$, were employed to forecast the values for the subsequent $64$ time steps. Detailed results are presented in Fig.~\ref{ablation_T}. SUMformer-AD also integrates a MHSA to extract temporal dependencies. However, unlike some prior models that may overfit to temporal noise, SUMformer-AD adeptly captures temporal information. While our model's performance is marginally suboptimal on TaxiBJ for input sequence lengths exceeding 64, there is a general trend of diminishing error. We believe this can be attributed to the SUMformer's temporal transformer operating within the temporal patch itself. By condensing long sequences into shorter ones via the temporal patch partition, it effectively counters the transformer's inherent limitations in grasping prolonged sequence correlations within time series.

\subsubsection{\textbf{Performance Across Different Patch Merge Windows}} Fig.~\ref{ablation_rwin} illustrates the relationship between the rise in $r_{win}$ and the number of patch merges, given that $N_{seg}$ remains constant at 16 and the network depth is fixed at 4. For uniformity in network depth, when $r_{win}$ is set to 2, patch merges are executed at every layer. With $r_{win}$ set to 4, patch merges occur at the first and third layers, while for $r_{win}$ at 8, they are only conducted at the initial layer. The findings indicate that an $r_{win}$ value of 2 yields the most favorable outcomes. The model without patch merge at $r_{win} = 1$ underperforms, emphasizing the critical role of patch merges in grasping multi-scale correlations. Conversely, a rise in $r_{win}$ causes a decline in performance when $r_{win} > 2$, suggesting the model's essentiality to discern small-scale nuances.

\subsubsection{\textbf{Performance with Different Spatial Patch Size for SUMformer-ViT}} To further validate that the multivariate time series perspective is better suited for urban mobility data, Fig.~\ref{ablation_lspatial} displays the RMSE values in relation to various spatial patch sizes. 
For simplicity, we present the results of SUMformer-AD in Fig.~\ref{ablation_lspatial} when the spatial patch size equals 1. For the TaxiBJ dataset, as the \( L_{{spatial}} \) value increases from 1 to 2, the RMSE significantly climbs. However, as the \( L_{{spatial}} \) continues to grow, the RMSE mildly declines, peaking at \( L_{{spatial}} = 4 \) and then slightly dropping at \( L_{{spatial}} = 8 \). A similar trend is observed for the Chengdu dataset. This supports our hypothesis. Although setting the patch size to 8 introduces more larger range correlations during token embedding computation, its performance is still not as good as when the patch size is 1. This suggests treating the data of each grid as an independent variable and then extracting downstream spatial correlations is a better approach for the grid-based urban mobility data.

\subsection{Attention Score Visualization}

\begin{figure}[htp]
\centering    
{\includegraphics[width=0.45\textwidth]{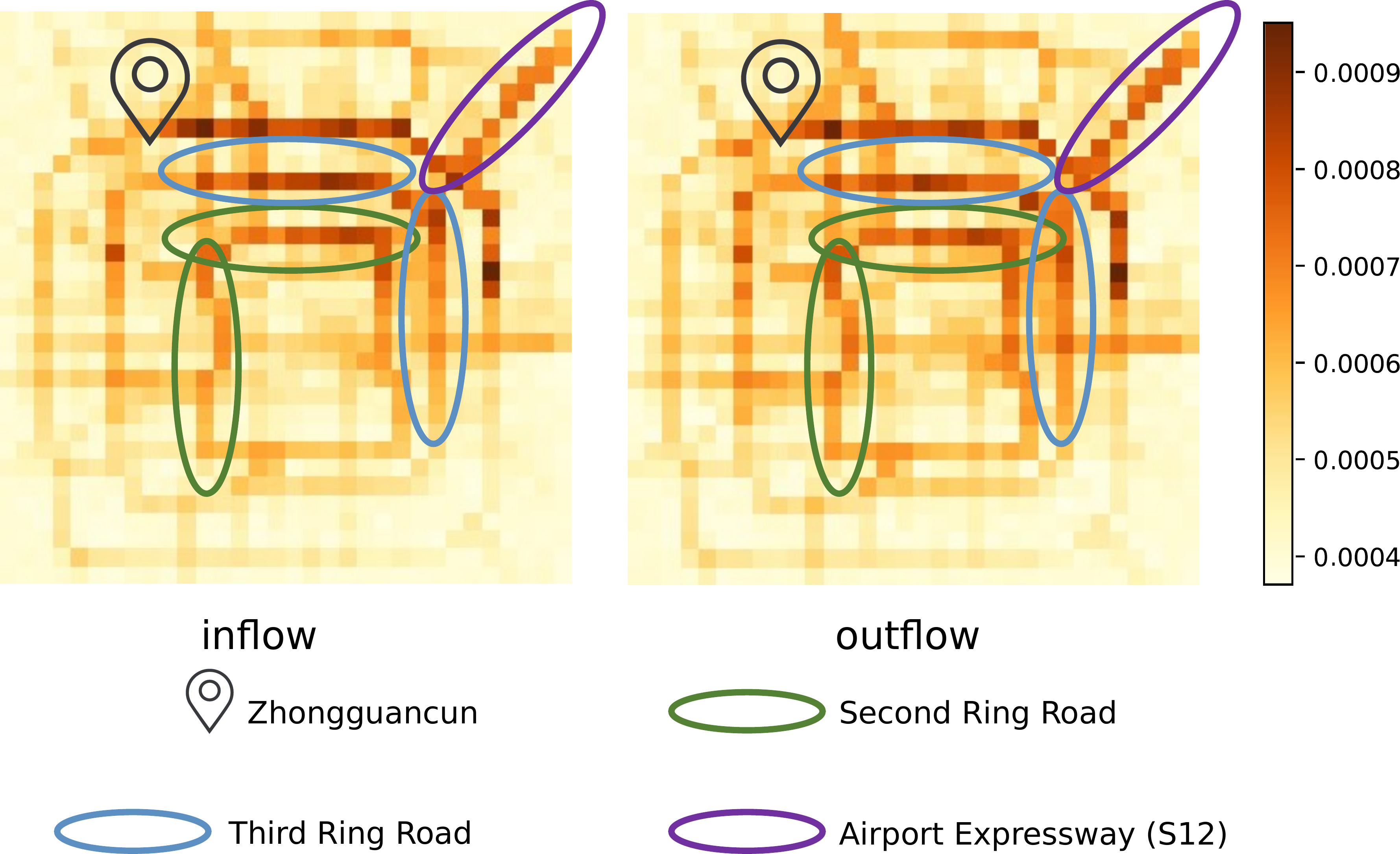}}
\caption{The attention score reflects the relationship between the inflow to Zhongguancun in the Haidian District and the in/out flow to various other regions throughout Beijing.}
\label{attention_region}
\end{figure}

%The core component of the SUMformer that captures dependencies between series is its self-attention layers, which calculate attention scores between each pair of points. To demonstrate how the SUMformer discerns these inter-series dependencies from urban mobility data, we present the attention scores between the inflow of the Zhongguancun (an area surrounding by universities and CBDs) area in relation to other regions. Fig.~\ref{Param_size} displays these raw attention scores. We have selected SUMformer-AF for this illustration due to its suitability for clear visualization. We observe that the super-multivariate perspective enables us to identify point-to-point relationships, which is unachievable with ViT and CNN architectures. The key takeaway from the figure is that the attention map of SUMformer can automatically identify significant areas, such as the main ring roads of Beijing, which are crucial for zhongguanchun's inflow and outflow. This suggests that the model is effectively learning and highlighting areas of traffic concentration without manual input.
 To elucidate the SUMformer's proficiency in unraveling the inter-series correlations, we turn our focus to the attention scores that pertain to the inflow of the Zhongguancun area\textemdash a hub of academic institutions and central business districts. Fig.~\ref{attention_region} offers a raw glimpse into these scores. For the sake of clarity in our visual exposition, we have employed SUMformer-AF for visualization.

A cursory glance at the attention maps reveals a nuanced picture. The SUMformer, with its super-multivariate lens, brings to light the point-to-point relationships that remain elusive to other architectures like ViTs and CNNs. The principal revelation from the figure is the attention map's innate capacity to autonomously spotlight key areas, the principal ring roads enveloping Beijing. These roads are not just mere strips of asphalt; they are the lifeblood of Zhongguancun's inflow, crucial arteries that dictate the pulse of traffic within the city. The attention map's ability to isolate these areas for scrutiny, absent any manual guidance, is indicative of the SUMformer's deep learning prowess. It identifies and accentuates areas where traffic congregates, providing invaluable insights into urban mobility patterns. 

%We visualize the attention scores both pre-softmax operation and the final scores. We observe that the attention distribution is quite sparse. The expansive multivariate perspective enables us to identify sparse point-to-point relationships, something that is unachievable with ViT and CNN architectures.
% 同时我们观察到，北京其他区域的inflow和outflow对zhongguancun的inflow流量影响存在细微区别。例如北二环与北三环的outflow相比较inflow，对zhongguancun的交通流量关联更大，这可能意味着北二环及北三环对zhongguancun的正向车流。zhongguancun作为一个商业发达的地区，对北京北部的交通流量吸引是比较make sense的。

%{\color{red}Additionally, there are subtle distinctions in the impact of inflow and outflow from other areas in Beijing on the inflow of Zhongguancun. For instance, when comparing the outflow relative to the inflow from the Second Ring Road, Third Ring Road as well as S12, a stronger correlation is observed with the inflow into Zhongguancun. This could imply that the model, based on historical information, has inferred that in the upcoming future time periods, the Second Ring Road, Third Ring Road and S12 will contribute a net inflow of traffic to Zhongguancun. Given Zhongguancun's status as a commercially developed area, its attractiveness to traffic flow from the northern Beijing aligns sensibly.}

\section{Conclusion}
In this study, we note that widely used CNNs and ViTs in video prediction architectures fall short in capturing crucial representations, as well as spatial and cross-channel correlations, essential for long-term grid-based urban mobility forecasting. To address this, we introduce SUMformer, a tailored backbone architecture specifically designed for grid-based urban mobility data, comprising three key components that focus on temporal dynamics, inter-series correlations, and frequency information. Our experimental results show that SUMformer delivers outstanding performance across three different datasets, demonstrating a remarkable versatility of the framework, as evidenced by our thorough analysis. 
Furthermore, our study reveals that grid-based urban mobility data represents a unique dataset within the domain of time series forecasting research, offering a more extensive array of data series than those typically encountered in existing datasets. 
In the future, we aim to extend the application of SUMformer to a broader range of video prediction tasks, including those involving AI4science datasets, which cover phenomena like weather patterns and fluid dynamics.

%在普适的视频数据集中，SUMformer的性能存疑。SUMformer其实更针对时序的语义建模，而不是空间语义建模。不过将视频数据集倒转一下，我们可以用TSB单独建模每一帧的图像，然后用ISSB建模时序，LFSB替换为2DFFT降噪模块，也许会取得较好效果。


\section*{Acknowledgment}

This work was supported by the Natural Science Foundation of Sichuan Province (No. 2023NSFSC1423), the Fundamental Research Funds for the Central Universities and the Natural Science Foundation of China (No. 62371324). We would like to thank Wang Kun from the University of Science and Technology of China for his helpful suggestions. We also acknowledge the generous contributions of dataset donors.



%\bibliographystyle{IEEEtran}


% \end{thebibliography}

\bibliographystyle{IEEEtran}
\bibliography{main}

\begin{thebibliography}{10}
\providecommand{\url}[1]{#1}
\csname url@samestyle\endcsname
\providecommand{\newblock}{\relax}
\providecommand{\bibinfo}[2]{#2}
\providecommand{\BIBentrySTDinterwordspacing}{\spaceskip=0pt\relax}
\providecommand{\BIBentryALTinterwordstretchfactor}{4}
\providecommand{\BIBentryALTinterwordspacing}{\spaceskip=\fontdimen2\font plus
\BIBentryALTinterwordstretchfactor\fontdimen3\font minus \fontdimen4\font\relax}
\providecommand{\BIBforeignlanguage}[2]{{%
\expandafter\ifx\csname l@#1\endcsname\relax
\typeout{** WARNING: IEEEtran.bst: No hyphenation pattern has been}%
\typeout{** loaded for the language `#1'. Using the pattern for}%
\typeout{** the default language instead.}%
\else
\language=\csname l@#1\endcsname
\fi
#2}}
\providecommand{\BIBdecl}{\relax}
\BIBdecl

\bibitem{wang2020deep}
S.~Wang, J.~Cao, and S.~Y. Philip, ``Deep learning for spatio-temporal data mining: A survey,'' \emph{IEEE transactions on knowledge and data engineering}, vol.~34, no.~8, pp. 3681--3700, 2020.

\bibitem{jin2023spatio}
G.~Jin, Y.~Liang, Y.~Fang, J.~Huang, J.~Zhang, and Y.~Zheng, ``Spatio-temporal graph neural networks for predictive learning in urban computing: A survey,'' \emph{arXiv preprint arXiv:2303.14483}, 2023.

\bibitem{bigdataset}
J.~Wang, J.~Jiang, W.~Jiang, C.~Han, and W.~X. Zhao, ``Towards efficient and comprehensive urban spatial-temporal prediction: A unified library and performance benchmark,'' \emph{arXiv preprint arXiv:2304.14343}, 2023.

\bibitem{zhang2017deep}
J.~Zhang, Y.~Zheng, and D.~Qi, ``Deep spatio-temporal residual networks for citywide crowd flows prediction,'' in \emph{Proceedings of the AAAI conference on artificial intelligence}, vol.~31, no.~1, 2017.

\bibitem{liu2018attentive}
L.~Liu, R.~Zhang, J.~Peng, G.~Li, B.~Du, and L.~Lin, ``Attentive crowd flow machines,'' in \emph{Proceedings of the 26th ACM international conference on Multimedia}, 2018, pp. 1553--1561.

\bibitem{lin2020preserving}
H.~Lin, R.~Bai, W.~Jia, X.~Yang, and Y.~You, ``Preserving dynamic attention for long-term spatial-temporal prediction,'' in \emph{Proceedings of the 26th ACM SIGKDD International Conference on Knowledge Discovery \& Data Mining}, 2020, pp. 36--46.

\bibitem{krizhevsky2012imagenet}
A.~Krizhevsky, I.~Sutskever, and G.~E. Hinton, ``Imagenet classification with deep convolutional neural networks,'' \emph{Advances in neural information processing systems}, vol.~25, 2012.

\bibitem{dosovitskiy2020image}
A.~Dosovitskiy, L.~Beyer, A.~Kolesnikov, D.~Weissenborn, X.~Zhai, T.~Unterthiner, M.~Dehghani, M.~Minderer, G.~Heigold, S.~Gelly \emph{et~al.}, ``An image is worth 16x16 words: Transformers for image recognition at scale,'' in \emph{International Conference on Learning Representations}, 2020.

\bibitem{arnab2021vivit}
A.~Arnab, M.~Dehghani, G.~Heigold, C.~Sun, M.~Lu{\v{c}}i{\'c}, and C.~Schmid, ``Vivit: A video vision transformer,'' in \emph{Proceedings of the IEEE/CVF international conference on computer vision}, 2021, pp. 6836--6846.

\bibitem{simvp}
Z.~Gao, C.~Tan, L.~Wu, and S.~Z. Li, ``Simvp: Simpler yet better video prediction,'' in \emph{Proceedings of the IEEE/CVF Conference on Computer Vision and Pattern Recognition}, 2022, pp. 3170--3180.

\bibitem{tang2023swinlstm}
S.~Tang, C.~Li, P.~Zhang, and R.~Tang, ``Swinlstm: Improving spatiotemporal prediction accuracy using swin transformer and lstm,'' in \emph{Proceedings of the IEEE/CVF International Conference on Computer Vision}, 2023, pp. 13\,470--13\,479.

\bibitem{guen2020disentangling}
V.~L. Guen and N.~Thome, ``Disentangling physical dynamics from unknown factors for unsupervised video prediction,'' in \emph{Proceedings of the IEEE/CVF Conference on Computer Vision and Pattern Recognition}, 2020, pp. 11\,474--11\,484.

\bibitem{yu2019efficient}
W.~Yu, Y.~Lu, S.~Easterbrook, and S.~Fidler, ``Efficient and information-preserving future frame prediction and beyond,'' in \emph{International Conference on Learning Representations}, 2019.

\bibitem{tau}
C.~Tan, Z.~Gao, L.~Wu, Y.~Xu, J.~Xia, S.~Li, and S.~Z. Li, ``Temporal attention unit: Towards efficient spatiotemporal predictive learning,'' in \emph{Proceedings of the IEEE/CVF Conference on Computer Vision and Pattern Recognition}, 2023, pp. 18\,770--18\,782.

\bibitem{convlstm}
Z.~Lin, M.~Li, Z.~Zheng, Y.~Cheng, and C.~Yuan, ``Self-attention convlstm for spatiotemporal prediction,'' in \emph{Proceedings of the AAAI conference on artificial intelligence}, vol.~34, no.~07, 2020, pp. 11\,531--11\,538.

\bibitem{e3dlstm}
Y.~Wang, L.~Jiang, M.-H. Yang, L.-J. Li, M.~Long, and L.~Fei-Fei, ``Eidetic 3d lstm: A model for video prediction and beyond,'' in \emph{International conference on learning representations}, 2018.

\bibitem{crevnet}
W.~Yu, Y.~Lu, S.~Easterbrook, and S.~Fidler, ``Crevnet: Conditionally reversible video prediction,'' \emph{arXiv preprint arXiv:1910.11577}, 2019.

\bibitem{vivit}
A.~Arnab, M.~Dehghani, G.~Heigold, C.~Sun, M.~Lu{\v{c}}i{\'c}, and C.~Schmid, ``Vivit: A video vision transformer,'' in \emph{Proceedings of the IEEE/CVF international conference on computer vision}, 2021, pp. 6836--6846.

\bibitem{videoswintransfomer}
Z.~Liu, J.~Ning, Y.~Cao, Y.~Wei, Z.~Zhang, S.~Lin, and H.~Hu, ``Video swin transformer,'' in \emph{Proceedings of the IEEE/CVF conference on computer vision and pattern recognition}, 2022, pp. 3202--3211.

\bibitem{ulyanov2018deep}
D.~Ulyanov, A.~Vedaldi, and V.~Lempitsky, ``Deep image prior,'' in \emph{Proceedings of the IEEE conference on computer vision and pattern recognition}, 2018, pp. 9446--9454.

\bibitem{d2021convit}
S.~d’Ascoli, H.~Touvron, M.~L. Leavitt, A.~S. Morcos, G.~Biroli, and L.~Sagun, ``Convit: Improving vision transformers with soft convolutional inductive biases,'' in \emph{International Conference on Machine Learning}.\hskip 1em plus 0.5em minus 0.4em\relax PMLR, 2021, pp. 2286--2296.

\bibitem{revisitingCNN}
Y.~Liang, K.~Ouyang, Y.~Wang, Y.~Liu, J.~Zhang, Y.~Zheng, and D.~S. Rosenblum, ``Revisiting convolutional neural networks for citywide crowd flow analytics,'' in \emph{Machine Learning and Knowledge Discovery in Databases: European Conference, ECML PKDD 2020, Ghent, Belgium, September 14--18, 2020, Proceedings, Part I}.\hskip 1em plus 0.5em minus 0.4em\relax Springer, 2021, pp. 578--594.

\bibitem{informer}
H.~Zhou, S.~Zhang, J.~Peng, S.~Zhang, J.~Li, H.~Xiong, and W.~Zhang, ``Informer: Beyond efficient transformer for long sequence time-series forecasting,'' in \emph{Proceedings of the AAAI conference on artificial intelligence}, vol.~35, no.~12, 2021, pp. 11\,106--11\,115.

\bibitem{zheng2014urban}
Y.~Zheng, L.~Capra, O.~Wolfson, and H.~Yang, ``Urban computing: concepts, methodologies, and applications,'' \emph{ACM Transactions on Intelligent Systems and Technology (TIST)}, vol.~5, no.~3, pp. 1--55, 2014.

\bibitem{boyce2015forecasting}
D.~E. Boyce and H.~C. Williams, \emph{Forecasting urban travel: Past, present and future}.\hskip 1em plus 0.5em minus 0.4em\relax Edward Elgar Publishing, 2015.

\bibitem{set_transformer}
J.~Lee, Y.~Lee, J.~Kim, A.~Kosiorek, S.~Choi, and Y.~W. Teh, ``Set transformer: A framework for attention-based permutation-invariant neural networks,'' in \emph{International conference on machine learning}.\hskip 1em plus 0.5em minus 0.4em\relax PMLR, 2019, pp. 3744--3753.

\bibitem{linformer}
S.~Wang, B.~Z. Li, M.~Khabsa, H.~Fang, and H.~Ma, ``Linformer: Self-attention with linear complexity,'' \emph{arXiv preprint arXiv:2006.04768}, 2020.

\bibitem{additive_attention}
C.~Wu, F.~Wu, T.~Qi, Y.~Huang, and X.~Xie, ``Fastformer: Additive attention can be all you need,'' \emph{arXiv preprint arXiv:2108.09084}, 2021.

\bibitem{swintransformer}
Z.~Liu, Y.~Lin, Y.~Cao, H.~Hu, Y.~Wei, Z.~Zhang, S.~Lin, and B.~Guo, ``Swin transformer: Hierarchical vision transformer using shifted windows,'' in \emph{Proceedings of the IEEE/CVF international conference on computer vision}, 2021, pp. 10\,012--10\,022.

\bibitem{lin2019deepstn+}
Z.~Lin, J.~Feng, Z.~Lu, Y.~Li, and D.~Jin, ``Deepstn+: Context-aware spatial-temporal neural network for crowd flow prediction in metropolis,'' in \emph{Proceedings of the AAAI conference on artificial intelligence}, vol.~33, no.~01, 2019, pp. 1020--1027.

\bibitem{predcnn}
Z.~Xu, Y.~Wang, M.~Long, J.~Wang, and M.~KLiss, ``Predcnn: Predictive learning with cascade convolutions.'' in \emph{IJCAI}, 2018, pp. 2940--2947.

\bibitem{zhang2020understanding}
H.~Zhang, Y.~Wu, H.~Tan, H.~Dong, F.~Ding, and B.~Ran, ``Understanding and modeling urban mobility dynamics via disentangled representation learning,'' \emph{IEEE Transactions on Intelligent Transportation Systems}, vol.~23, no.~3, pp. 2010--2020, 2020.

\bibitem{wang2017predrnn}
Y.~Wang, M.~Long, J.~Wang, Z.~Gao, and P.~S. Yu, ``Predrnn: Recurrent neural networks for predictive learning using spatiotemporal lstms,'' \emph{Advances in neural information processing systems}, vol.~30, 2017.

\bibitem{wang2018predrnn++}
Y.~Wang, Z.~Gao, M.~Long, J.~Wang, and S.~Y. Philip, ``Predrnn++: Towards a resolution of the deep-in-time dilemma in spatiotemporal predictive learning,'' in \emph{International Conference on Machine Learning}.\hskip 1em plus 0.5em minus 0.4em\relax PMLR, 2018, pp. 5123--5132.

\bibitem{wang2019memory}
Y.~Wang, J.~Zhang, H.~Zhu, M.~Long, J.~Wang, and P.~S. Yu, ``Memory in memory: A predictive neural network for learning higher-order non-stationarity from spatiotemporal dynamics,'' in \emph{Proceedings of the IEEE/CVF conference on computer vision and pattern recognition}, 2019, pp. 9154--9162.

\bibitem{vaswani2017attention}
A.~Vaswani, N.~Shazeer, N.~Parmar, J.~Uszkoreit, L.~Jones, A.~N. Gomez, {\L}.~Kaiser, and I.~Polosukhin, ``Attention is all you need,'' \emph{Advances in neural information processing systems}, vol.~30, 2017.

\bibitem{ViT}
A.~Dosovitskiy, L.~Beyer, A.~Kolesnikov, D.~Weissenborn, X.~Zhai, T.~Unterthiner, M.~Dehghani, M.~Minderer, G.~Heigold, S.~Gelly \emph{et~al.}, ``An image is worth 16x16 words: Transformers for image recognition at scale. arxiv 2020,'' \emph{arXiv preprint arXiv:2010.11929}, 2010.

\bibitem{mlpmixer}
I.~O. Tolstikhin, N.~Houlsby, A.~Kolesnikov, L.~Beyer, X.~Zhai, T.~Unterthiner, J.~Yung, A.~Steiner, D.~Keysers, J.~Uszkoreit \emph{et~al.}, ``Mlp-mixer: An all-mlp architecture for vision,'' \emph{Advances in neural information processing systems}, vol.~34, pp. 24\,261--24\,272, 2021.

\bibitem{mlpst}
Z.~Zhang, Z.~Huang, Z.~Hu, X.~Zhao, W.~Wang, Z.~Liu, J.~Zhang, S.~J. Qin, and H.~Zhao, ``Mlpst: Mlp is all you need for spatio-temporal prediction,'' \emph{arXiv preprint arXiv:2309.13363}, 2023.

\bibitem{pangu}
K.~Bi, L.~Xie, H.~Zhang, X.~Chen, X.~Gu, and Q.~Tian, ``Accurate medium-range global weather forecasting with 3d neural networks,'' \emph{Nature}, vol. 619, no. 7970, pp. 533--538, 2023.

\bibitem{autoformer}
H.~Wu, J.~Xu, J.~Wang, and M.~Long, ``Autoformer: Decomposition transformers with auto-correlation for long-term series forecasting,'' \emph{Advances in Neural Information Processing Systems}, vol.~34, pp. 22\,419--22\,430, 2021.

\bibitem{fedformer}
T.~Zhou, Z.~Ma, Q.~Wen, X.~Wang, L.~Sun, and R.~Jin, ``Fedformer: Frequency enhanced decomposed transformer for long-term series forecasting,'' in \emph{International Conference on Machine Learning}.\hskip 1em plus 0.5em minus 0.4em\relax PMLR, 2022, pp. 27\,268--27\,286.

\bibitem{sageformer}
Z.~Zhang, X.~Wang, and Y.~Gu, ``Sageformer: Series-aware graph-enhanced transformers for multivariate time series forecasting,'' \emph{arXiv preprint arXiv:2307.01616}, 2023.

\bibitem{TCN}
S.~Bai, J.~Z. Kolter, and V.~Koltun, ``An empirical evaluation of generic convolutional and recurrent networks for sequence modeling,'' \emph{arXiv preprint arXiv:1803.01271}, 2018.

\bibitem{Nlinear}
A.~Zeng, M.~Chen, L.~Zhang, and Q.~Xu, ``Are transformers effective for time series forecasting?'' in \emph{Proceedings of the AAAI conference on artificial intelligence}, vol.~37, no.~9, 2023, pp. 11\,121--11\,128.

\bibitem{patchtst}
Y.~Nie, N.~H. Nguyen, P.~Sinthong, and J.~Kalagnanam, ``A time series is worth 64 words: Long-term forecasting with transformers,'' \emph{arXiv preprint arXiv:2211.14730}, 2022.

\bibitem{Channel-Indepence}
L.~Han, H.-J. Ye, and D.-C. Zhan, ``The capacity and robustness trade-off: Revisiting the channel independent strategy for multivariate time series forecasting,'' \emph{arXiv preprint arXiv:2304.05206}, 2023.

\bibitem{crossformer}
Y.~Zhang and J.~Yan, ``Crossformer: Transformer utilizing cross-dimension dependency for multivariate time series forecasting,'' in \emph{The Eleventh International Conference on Learning Representations}, 2022.

\bibitem{timesnet}
H.~Wu, T.~Hu, Y.~Liu, H.~Zhou, J.~Wang, and M.~Long, ``Timesnet: Temporal 2d-variation modeling for general time series analysis,'' \emph{arXiv preprint arXiv:2210.02186}, 2022.

\bibitem{koopa}
Y.~Liu, C.~Li, J.~Wang, and M.~Long, ``Koopa: Learning non-stationary time series dynamics with koopman predictors,'' \emph{arXiv preprint arXiv:2305.18803}, 2023.

\bibitem{FNO}
Z.~Li, N.~Kovachki, K.~Azizzadenesheli, B.~Liu, K.~Bhattacharya, A.~Stuart, and A.~Anandkumar, ``Fourier neural operator for parametric partial differential equations,'' \emph{arXiv preprint arXiv:2010.08895}, 2020.

\bibitem{FNOimage}
W.~Johnny, H.~Brigido, M.~Ladeira, and J.~C.~F. Souza, ``Fourier neural operator for image classification,'' in \emph{2022 17th Iberian Conference on Information Systems and Technologies (CISTI)}.\hskip 1em plus 0.5em minus 0.4em\relax IEEE, 2022, pp. 1--6.

\bibitem{yang2023enhancing}
C.~Yang, X.~Chen, L.~Sun, H.~Yang, and Y.~Wu, ``Enhancing representation learning for periodic time series with floss: A frequency domain regularization approach,'' \emph{arXiv preprint arXiv:2308.01011}, 2023.

\bibitem{pathak2022fourcastnet}
J.~Pathak, S.~Subramanian, P.~Harrington, S.~Raja, A.~Chattopadhyay, M.~Mardani, T.~Kurth, D.~Hall, Z.~Li, K.~Azizzadenesheli \emph{et~al.}, ``Fourcastnet: A global data-driven high-resolution weather model using adaptive fourier neural operators,'' \emph{arXiv preprint arXiv:2202.11214}, 2022.

\bibitem{woo2022cost}
G.~Woo, C.~Liu, D.~Sahoo, A.~Kumar, and S.~Hoi, ``Cost: Contrastive learning of disentangled seasonal-trend representations for time series forecasting,'' \emph{arXiv preprint arXiv:2202.01575}, 2022.

\bibitem{liang2021fine}
Y.~Liang, K.~Ouyang, J.~Sun, Y.~Wang, J.~Zhang, Y.~Zheng, D.~Rosenblum, and R.~Zimmermann, ``Fine-grained urban flow prediction,'' in \emph{Proceedings of the Web Conference 2021}, 2021, pp. 1833--1845.

\bibitem{gao2022earthformer}
Z.~Gao, X.~Shi, H.~Wang, Y.~Zhu, Y.~B. Wang, M.~Li, and D.-Y. Yeung, ``Earthformer: Exploring space-time transformers for earth system forecasting,'' \emph{Advances in Neural Information Processing Systems}, vol.~35, pp. 25\,390--25\,403, 2022.

\bibitem{schlapfer2021universal}
M.~Schl{\"a}pfer, L.~Dong, K.~O’Keeffe, P.~Santi, M.~Szell, H.~Salat, S.~Anklesaria, M.~Vazifeh, C.~Ratti, and G.~B. West, ``The universal visitation law of human mobility,'' \emph{Nature}, vol. 593, no. 7860, pp. 522--527, 2021.

\bibitem{wang2023anti}
X.~Wang and L.~Sun, ``Anti-circulant dynamic mode decomposition with sparsity-promoting for highway traffic dynamics analysis,'' \emph{Transportation Research Part C: Emerging Technologies}, vol. 153, p. 104178, 2023.

\end{thebibliography}

\end{document}